\documentclass[letterpaper, 10 pt, conference]{ieeeconf}  %

\IEEEoverridecommandlockouts                              %

\usepackage[T1]{fontenc}    %
\usepackage{graphics}
\usepackage{multirow}
\usepackage{url}
\usepackage{xparse}
\usepackage{amssymb}
\usepackage{xspace}
\usepackage{nicefrac}
\usepackage{amsmath}
\usepackage{float}
\usepackage{amsthm}
\usepackage{mathtools}
\usepackage{amsfonts}
\usepackage{enumerate}
\usepackage{setspace}
\usepackage{tabularx}
\usepackage{wrapfig}
\usepackage{xfrac}
\usepackage{tikz}
\usepackage[most]{tcolorbox}
\usepackage{graphicx}
\usepackage{microtype}
\microtypesetup{nopatch=footnote} %
\usepackage{booktabs} %
\usepackage[colorlinks,allcolors=blue]{hyperref}

\usepackage{paralist}
\usepackage{caption}
\usepackage{subcaption}
\usepackage{colortbl}
\usepackage{lmodern}

\usepackage{balance}
\usepackage{sidecap}
\usepackage[capitalize,noabbrev]{cleveref}

\usepackage[belowskip=1pt,aboveskip=5pt,font=small]{caption}
\usepackage[belowskip=0pt,aboveskip=3pt,font=small]{subcaption}

\usepackage[all=normal,paragraphs=tight,floats=tight]{savetrees}
\usepackage{pdflscape}

\newcommand{\stopac}{\textsc{StopAction}\xspace}

\newcommand{\simpred}{\textit{Sim Predictivity}\xspace}
\newcommand{\simtransferpred}{\textit{Sim2Real Transfer}\xspace}
\usepackage[normalem]{ulem}
\newcommand{\MARFIFTH}[1]{\textcolor{green}{#1}\xspace}

\title{\LARGE \bf
What Do We Learn from a Large-Scale Study of Pre-Trained Visual Representations in Sim and Real Environments?
}

\author{Sneha Silwal$^{1*}$, Karmesh Yadav$^{2*}$, Tingfan Wu$^{1*}$, Jay Vakil$^{1*}$, Arjun Majumdar$^{2*}$, Sergio Arnaud$^{1*}$, \\Claire Chen$^{3}$, Vincent-Pierre Berges$^{1}$, Dhruv Batra$^{1,2}$, Aravind Rajeswaran$^{1}$, \\ Mrinal Kalakrishnan$^{1}$, Franziska Meier$^{1\dag}$ and Oleksandr Maksymets$^{1\dag}$%
\thanks{\footnotesize*Equal Contribution \; \; \footnotesize$\dag$Equal Contribution}%
\thanks{\footnotesize$^{1}$Meta AI
        {\texttt{\{ssilwal,tingfan,jayvakil,sergioarnaud, vincentpierre,aravraj,mrinal,fmeier,maksymets\}@meta.com}}}%
\thanks{\footnotesize$^{2}$Georgia Institute of Technology
        {\footnotesize\texttt{\{kyadav32,arjun.majumdar, dbatra\}}@gatech.edu}}%
\thanks{\footnotesize$^{3}$Stanford University
        {\footnotesize\texttt{clairech}@stanford.edu}}%
}

\begin{document}

\maketitle
\thispagestyle{empty}
\pagestyle{empty}

\begingroup
\let\clearpage\relax
\begin{abstract}
We present a large empirical investigation on the use of pre-trained visual representations (PVRs) for training downstream policies that execute real-world tasks. Our study  involves five different PVRs, each trained for five distinct manipulation or indoor navigation tasks. We performed this evaluation using three different robots and two different policy learning paradigms. From this effort, we can arrive at three insights: 1) the performance trends of PVRs in the simulation are generally indicative of their trends in the real world, 2) the use of PVRs enables a first-of-its-kind result with indoor ImageNav (zero-shot transfer to a held-out scene in the real world), and 3) the benefits from variations in PVRs, primarily data-augmentation and fine-tuning, also transfer to the real-world performance. See 
\href{http://pvrs-sim2real.github.io/}
{project website}\footnote{{http://pvrs-sim2real.github.io/}} for additional details and visuals. 
\end{abstract}

\section{INTRODUCTION}
\label{sec: intro}

The design of \emph{pre-trained visual representations} (PVRs) for sensorimotor 
control \cite{khandelwal2021simple, Parisi2022-PVR, yadav_ovrl, Nair_r3m_2022, Ma2022-VIP, Radosavovic_MVP_2022,yadav2023ovrl,cortex-vc1-2023} has been a promising development towards general-purpose visual perception for robotics, overcoming the problem of limited data for any one task and addressing the ability to generalize to new scenes.  Today this involves training neural networks (on internet-scale data) that embed camera images into a visual feature space that can support policy learning for a variety of tasks (locomotion, navigation, static and mobile manipulation). To measure the potential of PVRs on being general purpose vision backbones -- for a diverse set of Embodied AI and robotics tasks -- we have to evaluate them on a wide range of robotics tasks. Yet doing so on hardware is impractical if not infeasible for most researchers in the community. As a result, past studies on the effectiveness of PVRs have either focused on broad systematic analyses in simulation \cite{cortex-vc1-2023} or narrow small-scale experiments on hardware \cite{Nair_r3m_2022, Ma2022-VIP, Radosavovic_MVP_2022}. 

Simulation lends itself to broad and diverse evaluations, which are needed to evaluate the generalization capabilities of PVRs. Yet it is unclear how much of the results from simulation carries over to the real world where we would hope to deploy PVRs. This gap in our collective scientific knowledge is best captured by the following research question -- \textit{Can we use simulation to evaluate and benchmark PVRs and carry over the results to hardware? In other words, is the performance of PVRs in simulation predictive of their performance in the real world?}

To answer this question, we conducted the largest empirical study of PVRs in simulation and the real world to date: involving 5 PVRs (R3M~\cite{Nair_r3m_2022}, CLIP~\cite{radford2021learning}, MVP~\cite{Radosavovic_MVP_2022}, and two variants of VC-1~\cite{cortex-vc1-2023}), 3 different robot platforms (TriFinger, Franka Arm and Stretch), 2 policy-learning paradigms (imitation and reinforcement learning), and 5 distinct tasks (pushing a cube, picking up a bottle, opening a drawer, reaching targets, and image-goal navigation (ImageNav)). For each task, we have a matching simulation and hardware setup - such that we can compare PVRs performance in the simulated setting and the real-world counterpart. Our empirical study comprised a total of 348 experiments and \textbf{\emph{over 110 hours of robot experimentation}} on hardware. To ensure statistical significance, all experiments (except ImageNav) were conducted using three random seeds for training policies, enabling us to identify common trends and exceptions.
\begin{figure}[t]
    \centering
    \includegraphics[width=\columnwidth]{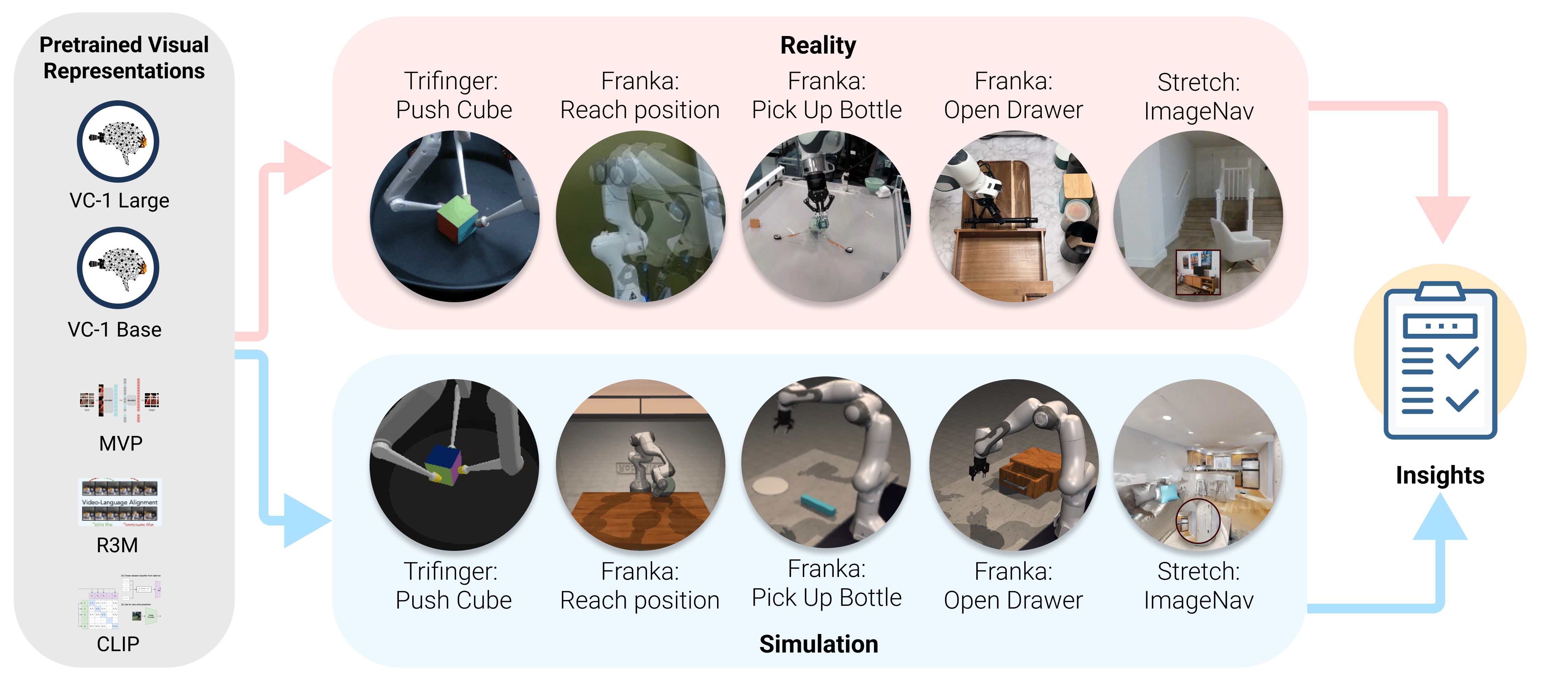}
    \caption{We conducted 348 experiments with PVRs on five tasks (push cube, pick up bottle, open drawer, reach goal position, and image-goal navigation (ImageNav)), three robots (Trifinger, Franka, and Stretch), two learning paradigms (imitation and reinforcement learning), in sim and reality.}
    \vspace*{-2pt}
    \label{fig:teaser}
\end{figure}

Our main findings and contributions are: 
\vspace*{-1pt}
\begin{enumerate}[\hspace{0pt}1.]
    \item \textbf{\simpred of PVR evaluations on hardware.} First, we ask if the performance of PVR-based policies in simulation is indicative of trends in the real world, despite potential variations in absolute performance metrics. In prior work~\cite{kadian2020sim2real}, \simpred is presented as a measure of how well performance translates from experiments conducted entirely in simulation to those carried out in real-world settings. Our hope is that if trends in simulation and real-world settings match, we can iterate faster on research ideas in simulation with greater confidence. We evaluate \textit{`sim-predictivity'} by training a set of policies in simulation and comparing their performance in simulation at evaluation time with similar policies trained with real demonstrations evaluated on hardware. In our experiments, we find a remarkably high degree of \textit{`sim-predictivity'} (a correlation coefficient of $0.929$) after basic alignment between the simulation and real-world setups (e.g.~camera placement, checkpoint selection schemes, etc.). This suggests that recent progress in training PVRs is indeed materializing into broadly applicable real-world gains and affirms the value of simulation benchmarks for model selection, but practices such as reporting maximum validation split performance should be reconsidered.
    \item \textbf{\simtransferpred of PVR-based policies.} We do the same analysis of \textit{`sim-predictivity'} with policies trained on demonstrations collected in simulation and evaluated on the real robot (addressing \simtransferpred). We find that in this setting the performance only transfers well for the ImageNav task. While most tasks do not exhibit \simtransferpred, an ImageNav agent is able to achieve a 90\% success rate in a zero-shot manner, making it a first result of this kind with regards to ImageNav policies trained on the HM3D dataset and evaluated in the real world.
    \item \textbf{Impact of Design Choices.} Finally, we study the effects of 1) model size, 2) fine-tuning the visual backbone, and 3) using data-augmentations on PVRs, with a focus on whether these results hold with our set of real-world experiments. Do the improvements we see in simulation also translate to hardware? We find the performance of most variations to be predictive of their real-world success. (\cref{sec:sim2real_imagenav}).
\end{enumerate}

\section{RELATED WORK}
\label{sec: related-work}

\textbf{Pre-trained Visual Representations.}
Inspired by the success of pre-trained representations for natural language processing and computer vision tasks, the robotics community has been exploring the use of PVRs to accelerate vision-based robotics, as opposed to the status quo of training models from scratch on in-domain data.
\cite{Parisi_unsurprising_2022} trains control policies with PVRs trained on large-scale computer vision datasets and show that in many cases, these policies are competitive or outperform policies trained with ground-truth state. \cite{Nair_r3m_2022}, \cite{Ma2022-VIP}, and \cite{Radosavovic_MVP_2022}, introduced the R3M, VIP, and MVP models respectively, all of which target manipulation tasks and train representations on egocentric video data. \cite{cortex-vc1-2023} introduced the VC-1 model, which was trained on egocentric data as well as web images. They also introduced a benchmark called CortexBench, consisting of a range of 17 different control tasks including locomotion, navigation, and mobile manipulation, however, all of these tasks were in simulation. Our work does not contribute a new PVR; instead we evaluate 5 previously published PVRs
 on a range of corresponding simulation and real-world tasks. 

\textbf{Sim2Real Transfer.} 
Compared to other domains, there is a dire lack of large, diverse, real-world robotics datasets. In this context, simulation is a highly appealing source of potentially unlimited data, but the use of this data to improve real-world performance can be challenging due to the domain gap between simulation and reality.
There have been a number of approaches which attempt to bridge this domain gap~\cite{kaspar2020sim2real,rao2020rl,ho2021retinagan,james2019sim,gervet2022navigating,mahler2017dex,sundermeyer2021contact}.
In particular, for image-based navigation tasks, \cite{gervet2022navigating} has shown that modular approaches using semantics significantly outperform end-to-end RL approaches when transferred to the real world. \cite{krantz2023navigating} arrives at similar findings, also demonstrating poor performance of an end-to-end network with a prior PVR. In contrast, in this work we show that VC-1 Large~\cite{cortex-vc1-2023}, which is pre-trained on diverse real-world data (including indoor environments), can be fine-tuned with RL in simulation to achieve a 90\% success rate on ImageNav in the real world.

\cite{Trauble_the-role-of_2022} has also explored the role of PVRs in training policies for Sim2Real transfer on robot tasks. However, while they use a PVR trained with visual data from a specific simulated robot setup, and apply it to the same robot in the real world, our study focuses on PVRs trained on out-of-domain real-world data, analyzing their applicability to multiple different robot platforms and settings.

\textbf{Sim2Real Predictivity.} 
\cite{kadian2020sim2real} investigates Sim2Real predictivity (what we refer to as \textit{`sim-predictivity'}) in the context of a visual navigation task. They introduce the Sim2Real Correlation Coefficient (SRCC) metric, which we also use in this work. However, while they study Sim2Real predictivity for a single task in a single simulation environment, our work extends this to a broader set of simulation environments and tasks, and studies Sim2Real in the context of PVRs.

\section{SIMULATED AND REAL-WORLD MANIPULATION AND NAVIGATION TASKS}
\label{sec:tasks}

\begin{figure*}
\vspace{5pt}
\centering
\begin{tabular}{ccccc}
\hspace{-30pt}
\subfloat[]{\includegraphics[angle=0,trim=0 0 0 0, clip, width=0.33\columnwidth]{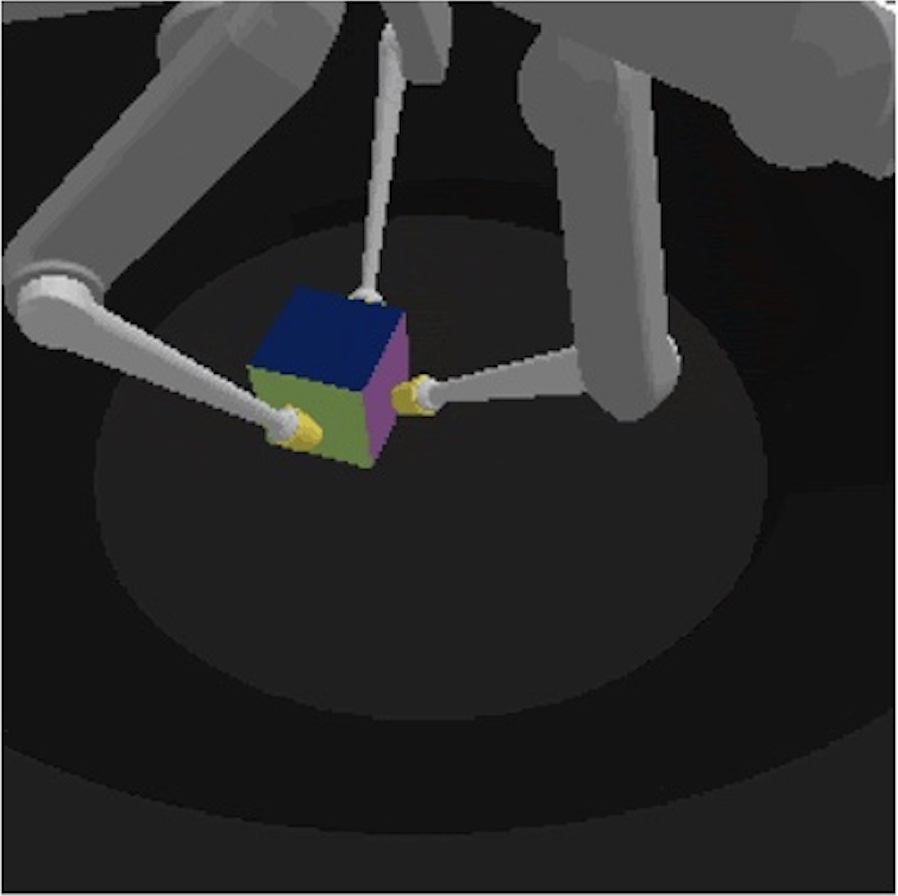}} & \hspace{-10pt}
\subfloat[]{\includegraphics[angle=0,trim=0 75 0 70, clip,width=0.33\columnwidth]{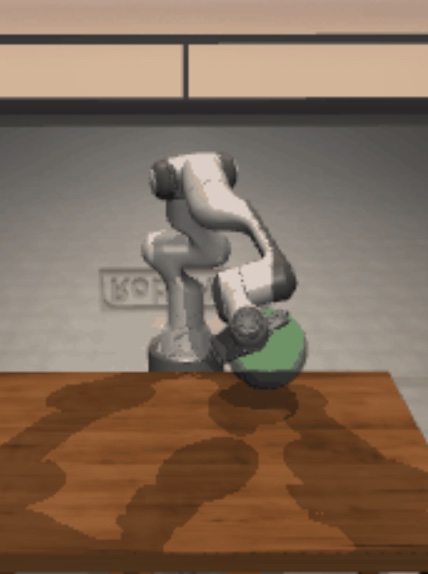}} & \hspace{-10pt}
\subfloat[]{\includegraphics[angle=0,trim=0 40 0 20, clip,width=0.33\columnwidth]{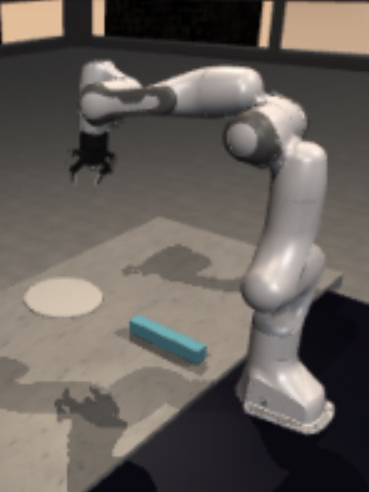}} & \hspace{-10pt}
\subfloat[]{\includegraphics[angle=0,trim=0 200 0 290, clip,width=0.33\columnwidth]{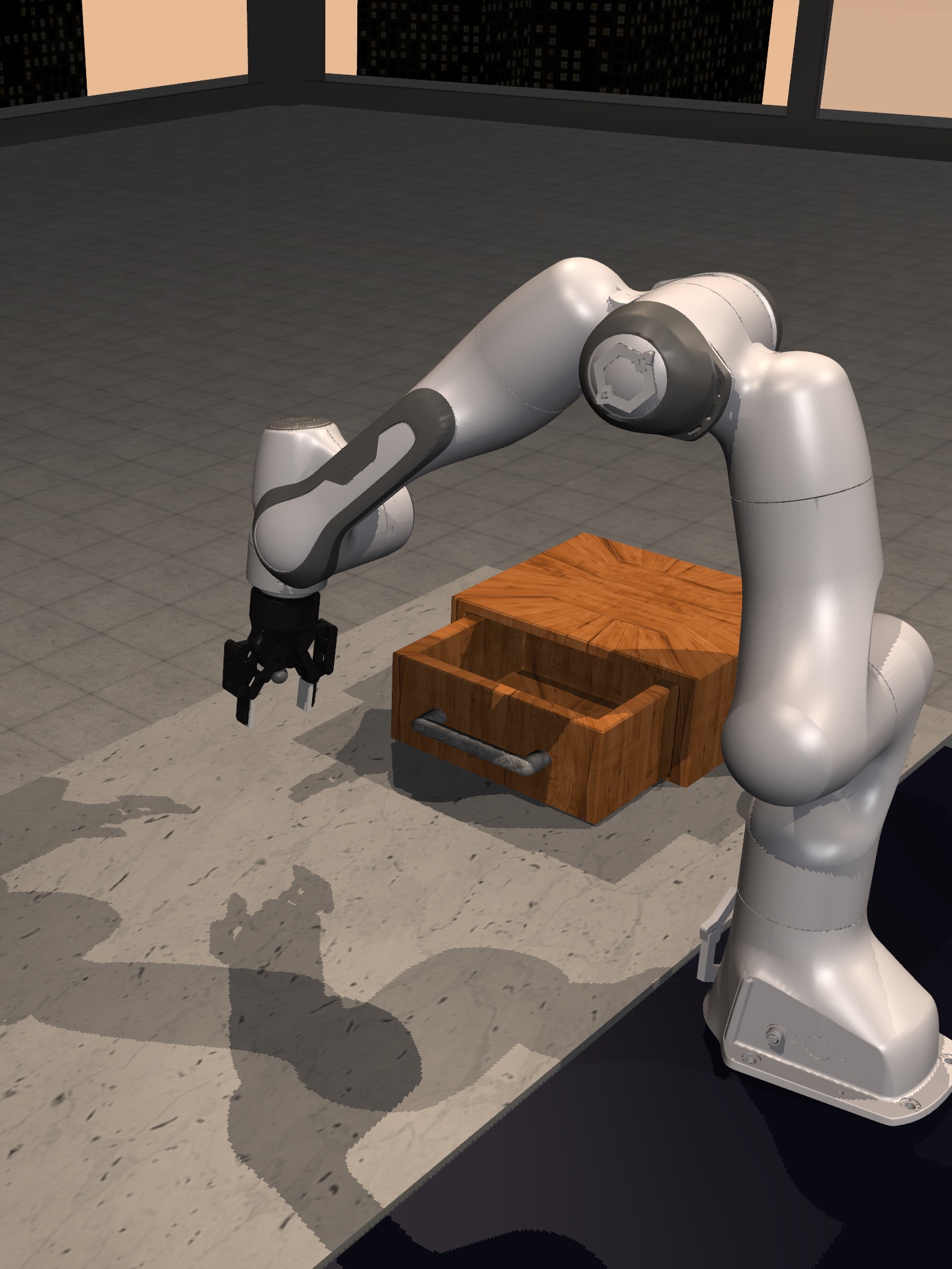}} & \hspace{-10pt}
\subfloat[]{\includegraphics[width=0.50\columnwidth]{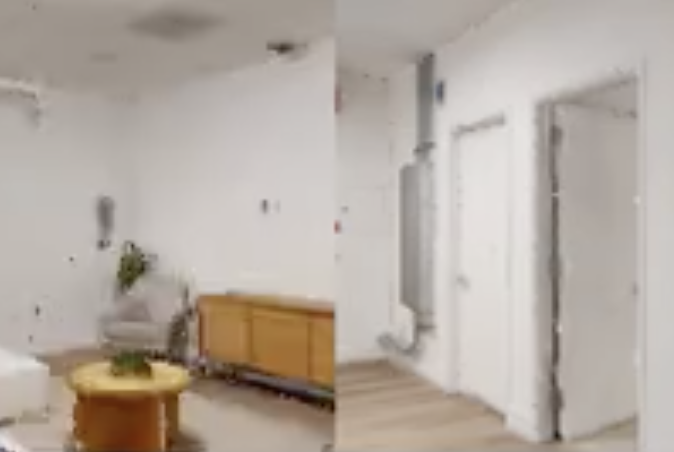}}\\
\hspace{-30pt}
\subfloat[Trifinger]{\includegraphics[angle=0,width=0.33\columnwidth]{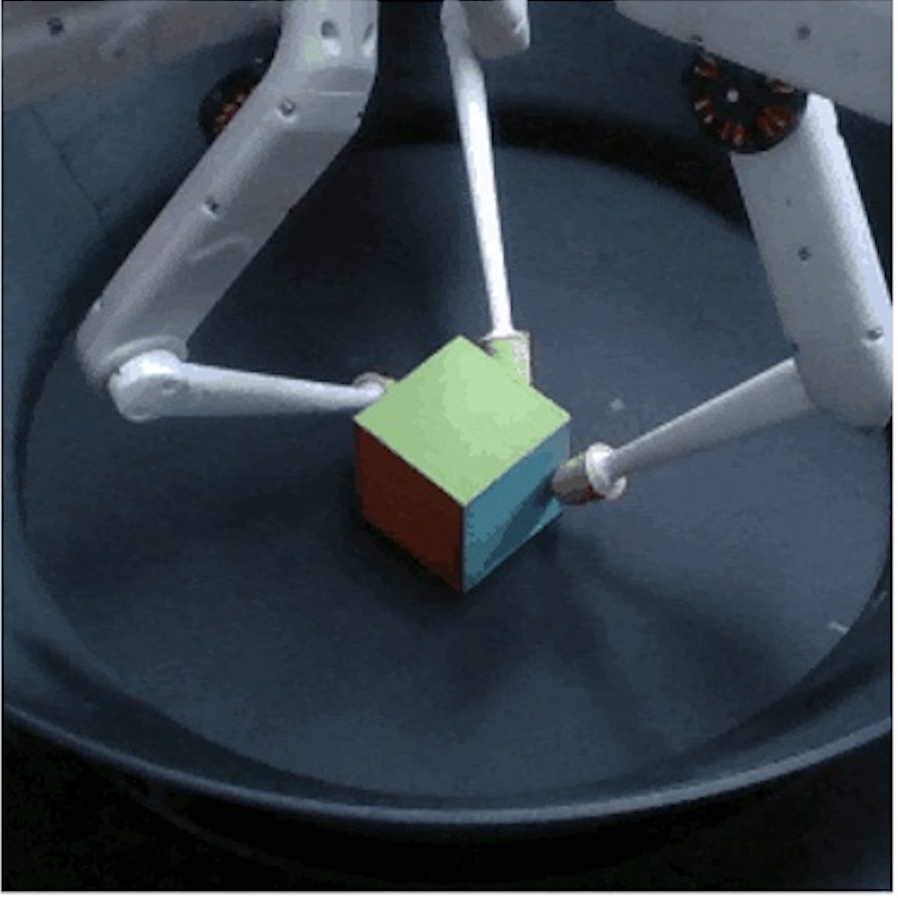}} & \hspace{-10pt}
\subfloat[Random Reach]{\includegraphics[angle=0, trim=0 0 0 73, clip, width=0.33\columnwidth]{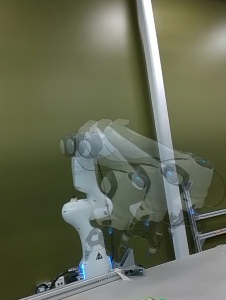}} & \hspace{-10pt}
\subfloat[Bottle Pickup]{\includegraphics[angle=0,trim=0 80 0 0, clip, width=0.33\columnwidth]{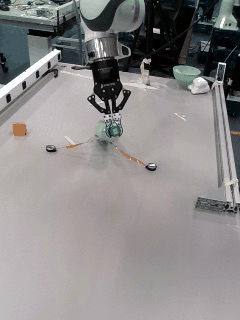}} & \hspace{-10pt}
\subfloat[Open Drawer]{\includegraphics[angle=-90,trim=120 00 120 0, clip,width=0.33\columnwidth]{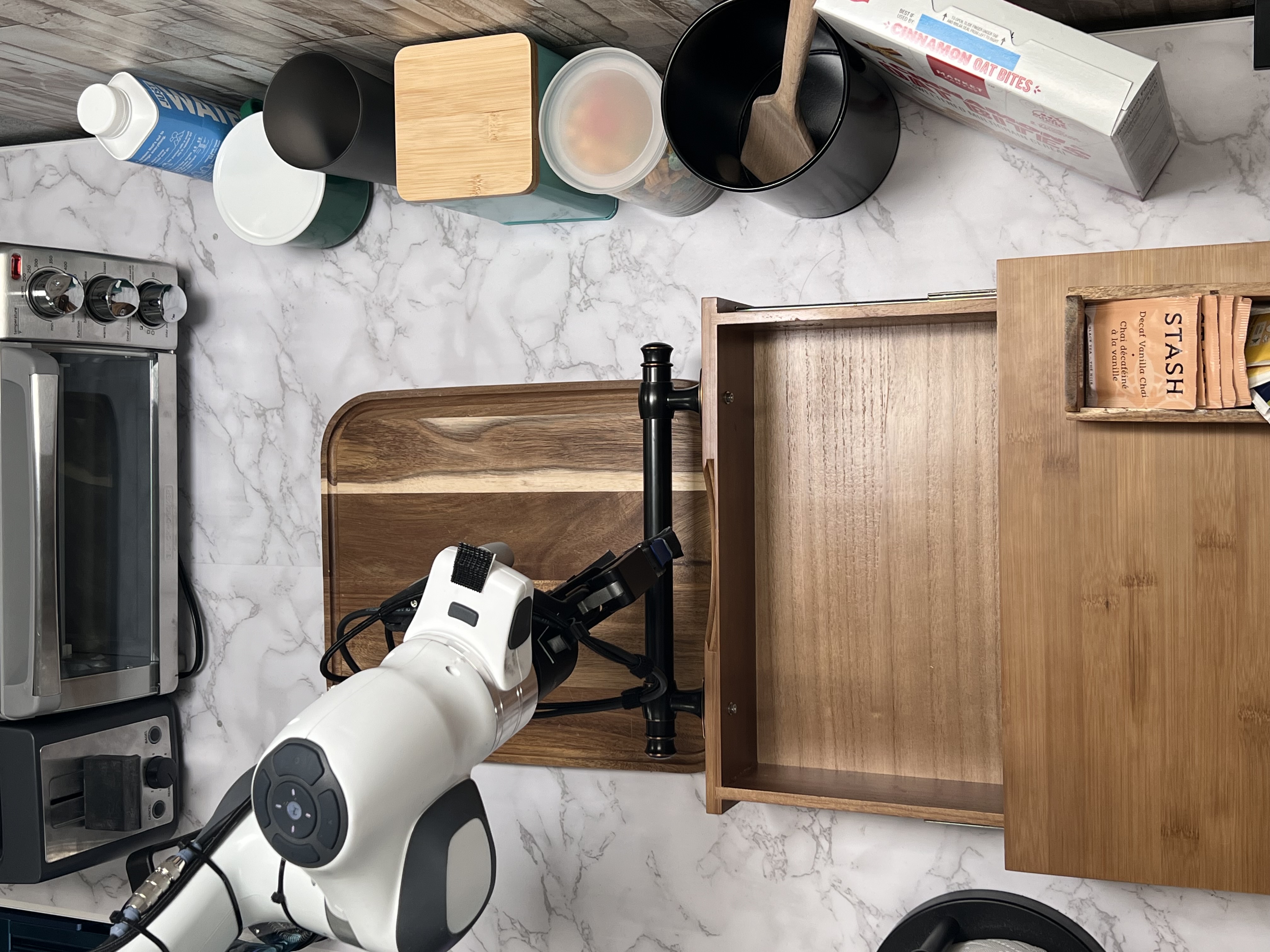}} & \hspace{-10pt}
\subfloat[ImageNav]{\includegraphics[width=0.5\columnwidth]{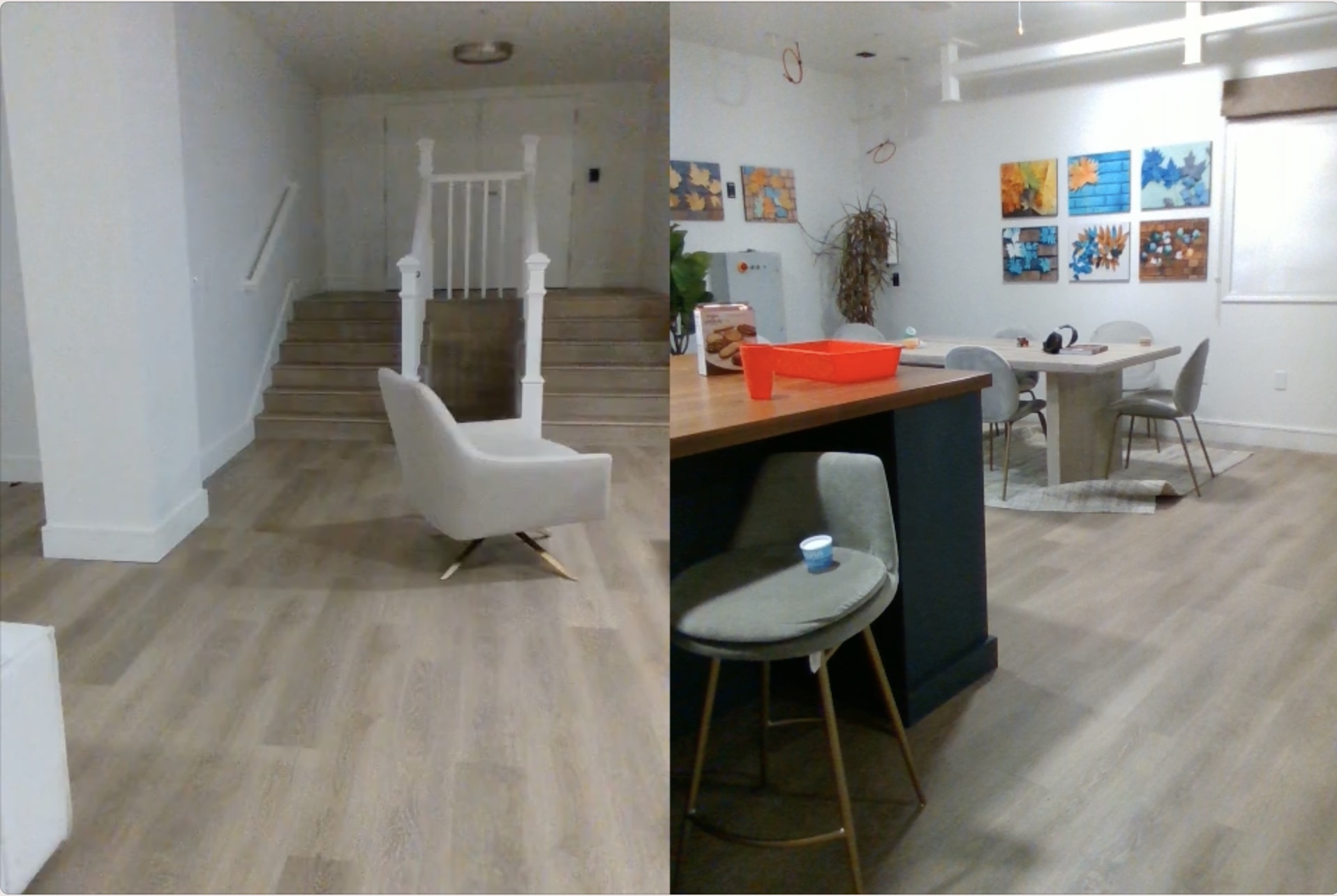}}
\end{tabular}
\caption{{\bf Top row:} Our 5 task in the simulation setting. {\bf Bottom row:} Corresponding tasks on hardware.}
\label{fig:franka-tasks-images}
\end{figure*}

In this section, we provide a brief description of the three robot platforms employed in our study and the associated tasks we set up to evaluate the PVRs both in simulation and real-world settings. In the simulation, we configure scenes to visually mimic the real world, including 3D models (rendering and collision meshes) of the robots, grippers, objects, and camera placement, but without rigorous system identification to match dynamics or camera extrinsic parameters.

\vspace*{-2pt}
\subsection{Planar Cube Manipulation with a Trifinger Robot via Behavior Cloning}
\label{sec:trifinger_details}

A TriFinger~\cite{Wuthrich_trifinger_2020} robot (as seen in \cref{fig:teaser}) consists of three 3-DoF fingers in a shared workspace and is used for fine-grained manipulation of objects. We consider a planar cube re-positioning task designed specifically to require visual perception and use behavior cloning (BC) for policy learning. 
For both simulation and reality, we collect 30 demonstrations each from an expert policy to train a BC policy for 500 epochs with a learning rate of $10^{-4}$. The architecture and training regime closely matches \cite{cortex-vc1-2023}.%
We used 3 seeds for training and evaluation and tested the policies on 12 different start and goal configurations, reporting the average success across seeds for both hardware and simulation experiments.

\vspace*{-2pt}
\subsection{Manipulation Tasks with a Franka Robot via Behavior Cloning}
\label{sec:franka_details}

We used a Franka Panda arm fitted with Festo adaptive fingers to solve three tasks: Reach Pos, Pick Up the Bottle, and Open the Drawer. The Reach Pos task is successful if the final position of the arm is within 5cm of the predetermined pose, which is a randomly sampled pose. The Pick Bottle task is successful if the arm picks up the bottle at the end of the episode. The Open Drawer task succeeds if the drawer is opened more than 50\% from its initial state.

\MARFIFTH{}{We collected demonstrations (30 for Reach Pos and Pick Up Bottle, 150 for Open Drawer from ~\cite{bharadhwaj2023roboagent}) in the real-world setup via human teleoperation using an Oculus Quest 2 controller \cite{kumar2023robohive}. The exact number of demonstrations was collected in simulation using a heuristic policy. We then trained policies using behavior cloning on both real-world and simulation demonstrations. The policies were trained for the same number of epochs  (200 for Reach Pos and Pick Up and 500 for Open Drawer) with three different seeds per task. We ran evaluations on the policies for ten episodes per task per seed and reported the average success.}

\vspace*{-2pt}
\subsection{Visual Navigation with a Stretch Robot via Large-Scale Reinforcement Learning}
For visual navigation, we use the Stretch robot, a mobile manipulator developed by Hello Robot. We pick the ImageNav task~\cite{zhu2017target}, where the agent must navigate to a goal location specified by an image in an unknown 3D environment. 
In contrast to the other tasks described above, we use reinforcement learning (RL) to train policies for ImageNav. Training is performed only in simulation, and the same policy is evaluated in simulation and the real world. For simulation, we leverage the Habitat~\cite{habitat19iccv} simulator and the HM3D scene dataset~\cite{ramakrishnan2021habitat}. We use all 800 HM3D scenes for training, and a simulated replica of an unseen real-world apartment for evaluation, testing the generalization capabilities of the visual encoder and policy. For real-world evaluation, we set up 10 different episodes with different start and goal positions in the unseen apartment. These episodes ranged from easy to hard, incorporating challenges such as multi-room navigation, disambiguation between similar goals using the background, and navigating around a kitchen island to reach the goal viewpoint. The robot's task is to reach the goal location while avoiding obstacles and minimizing collisions. 

For training the agents in the HM3D environments, we use 600M timesteps (25k updates) with 320 environments running in parallel. Each environment collects up to 64 frames of experience, followed by 2 PPO epochs utilizing 2 mini-batches. Unless otherwise specified, we use a learning rate of $2.5\times10^{-4}$ and update the parameters using the AdamW optimizer with a weight decay of $10^{-6}$. The reward functions are based on those presented in~\cite{al2022zero}, with success weighting $c_s = 5.0$, angle success weighting $c_a = 5.0$, goal radius $r_g = 1.0$, angle threshold $\theta_g = 25^{\circ}$, and slack penalty $\gamma = 0.01$. Performance is evaluated every 25M steps of training, and metrics are reported based on the highest success rate (SR) achieved on the validation set.

ImageNav performance was assessed using the following criteria:
\textbf{Success rate:} The proportion of successful episodes, where the robot reached the goal without significant collision and with a remaining distance of less than 1m.
\textbf{Number of steps:} The total number of steps taken by the robot to reach the goal in each episode.
\textbf{Distance to goal (m):} The Euclidean distance remaining between the robot and the goal at the conclusion of each episode.

\begin{table*} %
\vspace{9pt}

\caption{Success rate of policies on CortexBench and three hardware platforms (TriFinger, Franka, and Stretch) with results in reality (real) and simulation (sim). }

\label{tab:frozen_pvrs}
\centering
\resizebox{0.9\textwidth}{!}{
\begin{tabular}{l@{\hskip 4pt}lc|cccccccccccc}
\toprule
 & & \multicolumn{1}{c}{CortexBench} & \multicolumn{2}{c}{TriFinger} & \multicolumn{2}{c}{Franka} & \multicolumn{2}{c}{Franka} & \multicolumn{2}{c}{Franka} & \multicolumn{2}{c}{Stretch} & \multicolumn{2}{c}{Average}\\
\cmidrule(lr){3-3}
\cmidrule(lr){4-5}
\cmidrule(lr){6-7}
\cmidrule(lr){8-9}
\cmidrule(lr){10-11}
\cmidrule(lr){12-13}
\cmidrule(lr){14-15}
 & & All benchmarks & \multicolumn{2}{c}{Push cube} & \multicolumn{2}{c}{Reach pos.} & \multicolumn{2}{c}{Pick up bottle} & \multicolumn{2}{c}{Open drawer} & \multicolumn{2}{c}{ImageNav} & \multicolumn{2}{c}{All tasks}\\
\texttt{\#} & Method & (sim) & (sim) & (real) & (sim) & (real) & (sim) & (real) & (sim) & (real) & (sim) & (real) & (sim) & (real)\\ 
\midrule
\texttt{1} & R3M~\cite{Nair_r3m_2022}               &     58 &     31 &     34 & \bf 97 & \bf 100 & \bf 87 & \bf 87 &     37 &     37 &     25 &     20 &     56 &     56 \\
\texttt{2} & CLIP~\cite{radford2021learning}        &     57 &     31 &     38 &     80 &      80 &     77 &     63 &     40 &     33 &     39 &     20 &     54 &     47 \\
\texttt{3} & MVP~\cite{Radosavovic_MVP_2022}        &     68 & \bf 44 & \bf 46 &     90 &      90 &     70 &     50 &     50 &     27 & \bf 60 &     50 &     63 &     53 \\
\texttt{4} & VC-1 Base~\cite{cortex-vc1-2023}       &     66 &     40 &     37 & \bf 97 &      97 &     80 &     83 & \bf 57 & \bf 67 & \bf 61 &     60 & \bf 67 & \bf 69 \\
\texttt{5} & VC-1 Large~\cite{cortex-vc1-2023}      & \bf 69 &     41 &     38 & \bf 97 &      87 &     77 &     43 &     50 &     57 & \bf 60 & \bf 90 &     65 &     63 \\
\bottomrule

\end{tabular}
}
\vspace*{-5pt}
\end{table*}

\section{EXPERIMENTAL FINDINGS}
\label{sec:result}
In this section, we evaluate policies that use different PVRs on five tasks across three hardware platforms. In total, our real-world experiments required 110 hours of hands-on evaluation. Our experiments address the following questions:

\begin{compactenum}[\hspace{5pt}1.]
    \item How do recently released PVRs, designed specifically for robotics, perform across diverse simulated and real-world robotic tasks? (\cref{sec:eval_frozen_pvrs})

    \item How predictive are simulation evaluations of hardware evaluations when policies for real world experiments are trained from real world demonstrations and using frozen PVRs? How can we enhance the correlation between simulation and hardware results to improve simulation predictivity of real-world results?  (\cref{sec:sim2real-predictivity})

   \item  How does sim predictivity of PVR results change when we transfer policies from sim to real (Sim2Real transfer)? To what extent do policies trained with PVRs in simulation transfer to real-world scenarios? (\cref{sec:sim2real_imagenav})

    \item How do model size, fine-tuning, and data augmentations impact simulation predictivity? Can we utilize simulation to benchmark such variations? %
\end{compactenum}

\subsection{Evaluating Pre-Trained Visual Representations (PVRs) in Simulation and Reality}
\label{sec:eval_frozen_pvrs}

In this section, we study how various PVRs perform in both simulation and real-world experiments. We select five PVRs shown in~\cref{tab:frozen_pvrs} with success rates on the previously introduced simulation benchmark suite, CortexBench~\cite{cortex-vc1-2023}, ranging from 57\% to 69\%. We evaluate these PVRs on three platforms (TriFinger, Franka, and Stretch) in `frozen mode' -- i.e., the parameters of the PVR are \emph{not} updated during the policy learning stage. The CortexBench tasks span 17 different tasks, although the set does not include the exact same tasks examined in our study. While the Stretch ImageNav and Trifinger Push Cube tasks are the same, we use similar manipulation tasks to compare with our Franka tasks. We would expect the performance of this subset of CortexBench tasks to correspond with the performance in sim we observe. %

In~\cref{tab:frozen_pvrs}, we show results for both simulation and hardware evaluations. Except for the ImageNav task, all results in reality were achieved by training a policy on real robot demonstrations. We observe that MVP, R3M, VC-1 Base and VC-1 Large perform strongly for various tasks, and there is no one PVR that dominates across all tasks. In addition, each task’s success is defined differently, resulting in variation for the performance of a PVR across all the tasks. We consider a PVR to be stronger if its average success is higher, and here we see that VC-1 Base (row 4) has the highest average success rate of  69\%  on all five real-world robotics tasks.
CLIP (row 2) has the lowest average performance with 47\% success in reality. In general, these trends are similar (but not the same) as the trends on CortexBench, where VC-1 Base is the third-best PVR and CLIP is the lowest-ranked method.

The correlation between performance on CortexBench and our real-world evaluations is further studied in \cref{sec:sim2real-predictivity}. Below, we analyze the trends in these real-world results on individual tasks.

The strongest performing model differs by task: MVP for the Trifinger task, R3M and VC-1 Base, the two smallest models, for Franka tasks, and VC-1 Large for the ImageNav task on the Stretch. Unlike other tasks, the policies for ImageNav were trained with large-scale reinforcement learning and evaluated in a Sim2Real manner.
We hypothesize that VC-1 Large does not perform as well as VC-1 Base due to the limited amount of data our policies were trained with, and this would be consistent with the findings in\cite{cortex-vc1-2023}, in which scaling data and model size does not always lead to better performance.

\subsection{Sim Predictivity of hardware results when policies are trained on real demonstrations}
\label{sec:sim2real-predictivity}

In this section, we analyze the correlation between performance on CortexBench and performance on real robots and study how mirroring real-world experimental conditions in evaluations conducted in simulation can further improve \simpred (as measured by SRCC~\cite{kadian2020sim2real}). Specifically, we contrast the simulation benchmark proposed in~\cite{cortex-vc1-2023} (CortexBench) with our simulated evaluations from~\cref{sec:tasks}. 

For this analysis, we subselected the four tasks that have real-world demonstrations available (from the TriFinger and Franka platforms). We want to understand how well simulation performance translates to real-world performance when training on real-world data. The Stretch/ImageNav task was left out of this analysis since it is trained entirely in simulation and thus is a case of \textit{Sim2Real Transfer}, discussed further in \cref{sec:sim2real_imagenav}.

In~\cref{tab:frozen_pvrs}, we report the performance in simulations (sim) designed for the tasks studied in this work.  Their simulation settings are discussed in~\cref{sec:tasks} and were constructed differently from CortexBench to closely reflect real-world conditions in a few different ways: the number of demonstrations in sim were adjusted to match the number available in the real world, and the last training checkpoint is chosen instead of the best performing on a validation set, since doing that would be prohibitive in the real world.

\begin{figure}
    \vspace{5pt}
    \centering
    \includegraphics[width=0.35\columnwidth]{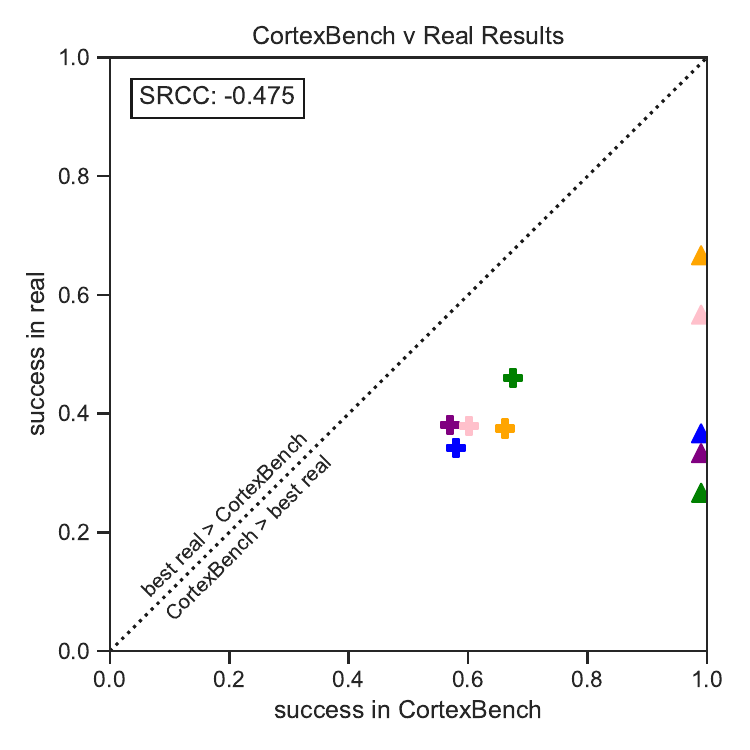}
    \includegraphics[width=0.35\columnwidth]{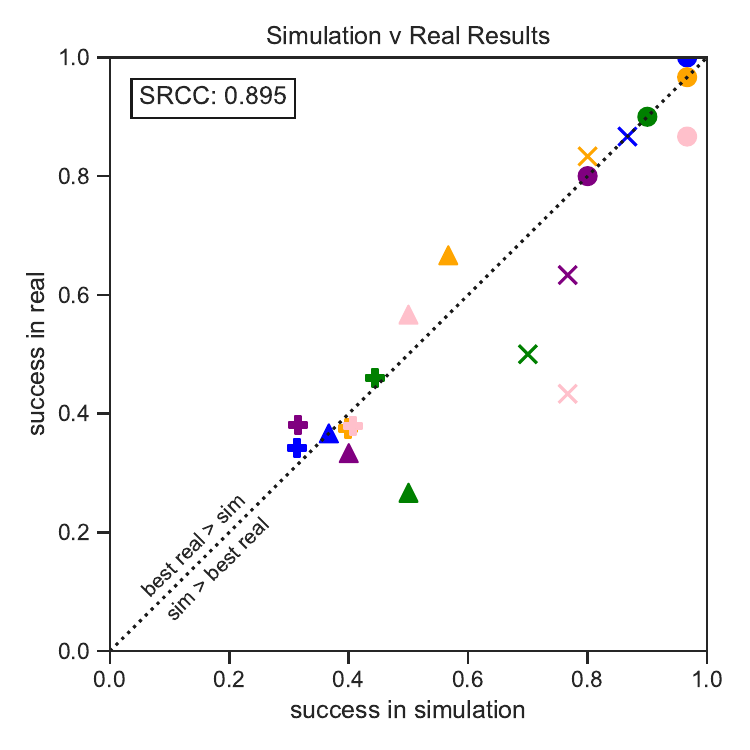}
    \raisebox{7mm}{    
        \includegraphics[width=0.15\columnwidth]{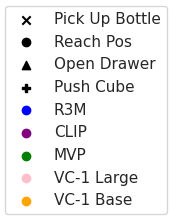}
    }
    \caption{Comparison of \simpred between CortexBench (left) and our simulation setting (right). Each data point represents a (model, task) tuple. Models and tasks are depicted by colors and symbols respectively, as shown in the legend.
}
    \label{fig:srcc_frozen_pvrs_averaged}
\end{figure}

\cref{fig:srcc_frozen_pvrs_averaged} shows the correlations between simulation and real-world performance, for both Cortexbench and our simulation settings. Analyzing the correlations between CortexBench and real-world performance results in an SRCC value of -0.475, while repeating the calculation with our simulation results in a substantially higher SRCC of 0.895. While this is not a surprising result, these results reinforce the importance of matching real-world settings in simulation in order to improve the predictability of real-world performance from results in simulation.

\begin{table}[h]
\caption{Zero-shot sim2real evaluations of randomly initialized ViT-Base model with finetuning \& augmentations (row 0) and pre-trained visual encoders (rows 1-5) for all tasks.}
\label{fig:sim2real_comparison_all_tasks}\centering
\resizebox{0.5\textwidth}{!}{
\begin{tabular}{l@{\hskip 4pt}lccccccccccc}
        \toprule
         & & \multicolumn{2}{c}{TriFinger} & \multicolumn{2}{c}{Franka} & \multicolumn{2}{c}{Franka} & \multicolumn{2}{c}{Franka} & \multicolumn{2}{c}{Stretch}\\
        \cmidrule(lr){3-4}
        \cmidrule(lr){5-6}
        \cmidrule(lr){7-8}
        \cmidrule(lr){9-10}
        \cmidrule(lr){11-12}
         & & \multicolumn{2}{c}{Push cube} & \multicolumn{2}{c}{Reach pos.} & \multicolumn{2}{c}{Pick up bottle} & \multicolumn{2}{c}{Open drawer} & \multicolumn{2}{c}{ImageNav} \\
        \texttt{\#} & Model & (real) & (sim) & (real) & (sim) & (real) & (sim) & (real) & (sim) & (real) & (sim) \\
        \midrule
        \texttt{0} & Scratch     & 8 & 21 & 0 & 27 & 0 & 11 & 30 & 37 & 10 & 35 \\
        \texttt{1} & R3M        & 11 & 31 & 0 & 97 & 0 & 87 & 10 & 37 & 20 & 25 \\
        \texttt{2} & CLIP       & 8 & 31 & 0 & 80 & 0 & 77 & 27 & 40 & 20 & 39 \\
        \texttt{3} & MVP        & 5 & 44 & 0 & 90 & 0 & 70 & 13 & 50 & 50 & 60 \\
        \texttt{4} & VC-1 Base  & 3 & 40 & 0 & 97 & 0 & 80 & 23 & 57 & 60 & 61 \\
        \texttt{5} & VC-1 Large & 2 & 41 & 0 & 97 & 0 & 77 & 23 & 50 & 90 & 60 \\
        \bottomrule
    \end{tabular}
}
\end{table}

\subsection{Effect of Sim2Real Policy Transfer on Simulation Predictivity}
\label{sec:sim2real_imagenav}

\begin{figure}[h]
    \vspace{5pt}
    \centering
    \includegraphics[width=0.48\columnwidth]{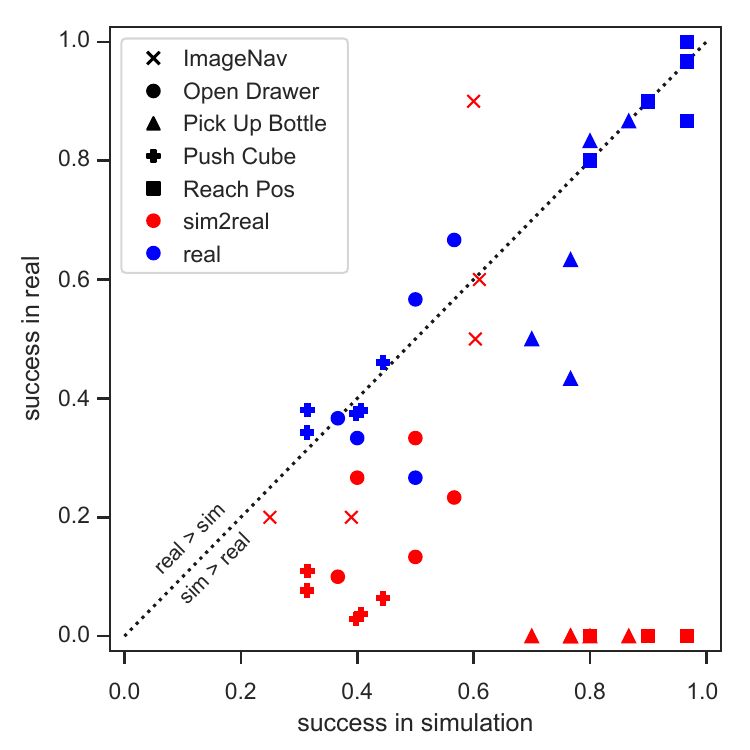}
    \caption{\simpred chart comparing correlations of sim performance to policies trained in the real world (blue), vs correlations of sim performance to policies trained in sim and transferred to hardware (red). Sim2Real transfer (red) is poor across the board for tasks that use few-shot imitation learning; as seen by the red points at the bottom of the plot. Transfer performance is substantially better on ImageNav (red cross markers), which is trained using large-scale reinforcement learning on simulated scenes.\label{fig:sim2real_comparison}}
\end{figure}

Having established that \simpred shows that reproducing setups in simulation and reality will yield similar levels of performance for the frozen PVRs, the natural next question is: how well do PVRs based policies perform when trained in simulation and transferred to the real world? We evaluated our simulation-trained policies obtained using 5 different PVRs on our 5 tasks and plotted the results in \cref{fig:sim2real_comparison}. The results suggest that for most tasks, specifically the ones trained using few-shot imitation learning, the performance achieved when running a simulation-trained policy in the real world can not be predicted by that in simulation, with most tasks' success metrics drop to near zero values (\cref{fig:sim2real_comparison_all_tasks}).

\begin{figure*}[h!]
    \vspace{5pt}
    \centering
    \includegraphics[width=0.32\textwidth]{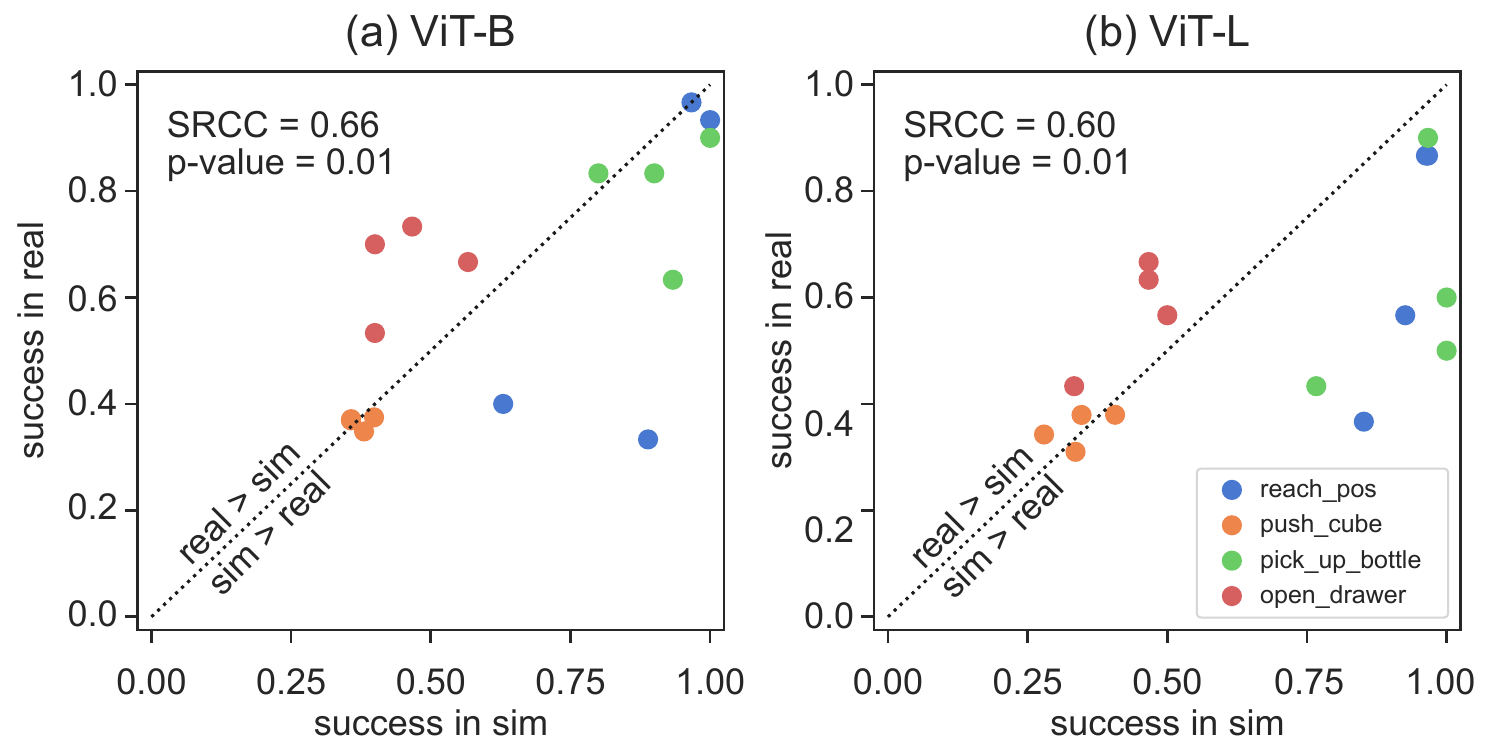}    \includegraphics[width=0.32\textwidth]{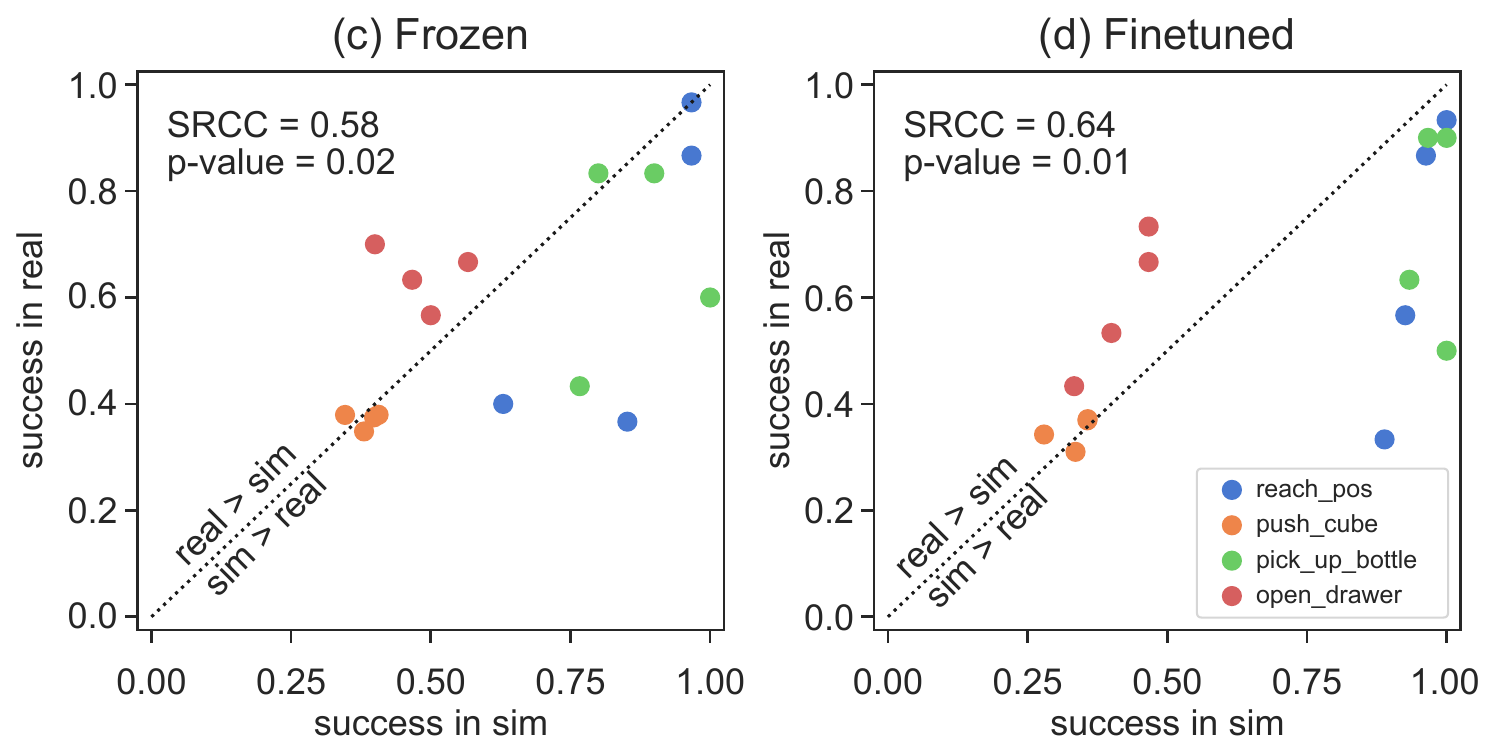}
    \includegraphics[width=0.32\textwidth]{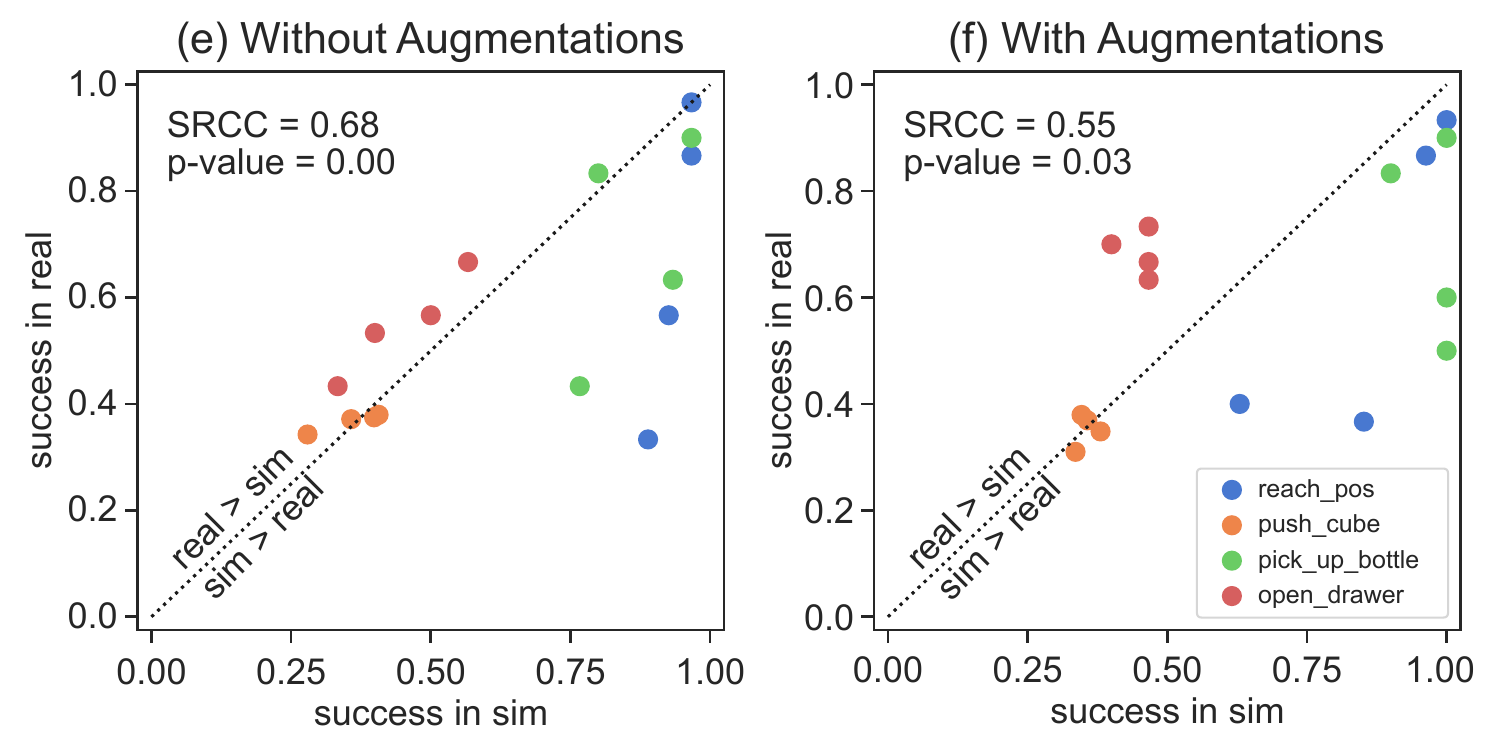}
    \caption{\simpred correlation plots analyzing the impact of different model variations of VC-1: (a, b) model size; (c, d) fine-tuning; (e, f) data augmentation.}
    \label{fig:vcvariationssrcc}
\end{figure*}

\begin{table}[!thb]
    \caption{Zero-shot Sim2Real generalization of policy with randomly initialized (row 0) or pre-trained visual representations (rows 1-5) on the ImageNav task with a Stretch robot.}
    \label{tab:imagenav_only}
    \centering
    \scalebox{0.75}{
        \setlength\tabcolsep{2.5pt} %
        
        \begin{tabular}{c@{\hskip 4pt}lcccccccccc}
        \toprule
        & & \multicolumn{2}{c}{Success (\%) $\uparrow$} & \multicolumn{2}{c}{Num Steps $\downarrow$} & \multicolumn{2}{c}{Dist-To-Goal $\downarrow$} & \multicolumn{2}{c}{Collisions $\downarrow$} \\
        \cmidrule(lr){3-4}
        \cmidrule(lr){5-6}
        \cmidrule(lr){7-8}
        \cmidrule(lr){9-10}
        \multicolumn{1}{c}{\texttt{\#}} & \multicolumn{1}{l}{Model} & (sim) & (real)  & (sim) & (real)  & (sim) & (real)  & (sim) & (real) \\
        \midrule
        \texttt{0} & Scratch    & 34.7     & 10.0     & 124.2    & 119.2  & 3.3     & 6.0     & 11.2    & 3.0     \\
        \texttt{1} & R3M        & 25.0     & 20.0     & \bf 71.0 & 81.7   & 3.4     & 4.2     & 5.1     & 1.7     \\
        \texttt{2} & CLIP       & 39.0     & 20.0     & 74.3     & 79.5   & 3.4     & 2.6     & \bf 2.5 & 0.6     \\
        \texttt{3} & MVP        & 60.3     & 50.0     & 97.0     & 114.8  & 2.2     & 1.8     & 4.2     & \bf 0.3 \\
        \texttt{4} & VC-1 Base  & \bf 61.0 & 60.0     & 101.0    & 109.7  & \bf 2.1 & 1.6     & 4.1     & 1.2     \\
        \texttt{5} & VC-1 Large & 60.0     & \bf 90.0 & 107.0    & 122.5  & 2.2     & \bf 0.8 & 3.9     & 1.0     \\
        \bottomrule
        \end{tabular}
    }
\end{table}

In contrast, the performance of frozen PVRs on the ImageNav task trained using large-scale RL is comparable to its performance in sim, which we may consider as a successful zero-shot Sim2Real transfer of results. In ~\cref{tab:imagenav_only}, we compare the performance of PVRs (rows 1-5) with a randomly initialized model that was fine-tuned with data augmentations for the ImageNav task (row 0). We find that in reality, the best model (VC-1 Large, row 5) far exceeds random-initialization performance by 80\% absolute (90\% vs. 10\%). This result strongly suggests that PVRs can play a crucial role in successful, zero-shot Sim2Real transfer. Achieving this result required changes to the original settings from CortexBench - particulary, changing the horizontal field of view on simulation to 42$^{\circ}$ to match that of the real robot.

\textit{Qualitative assessments for Sim2Real transfer on ImageNav:} We observed that \textbf{R3M} only completed the shorter episodes during real-world evaluation and frequently collided with obstacles (which resulted in episode termination), which reflects the behavior seen in simulation where the average number of collisions (5.1) exceeded other PVRs.

\textbf{CLIP} and \textbf{MVP} were both effective at avoiding obstacles (colliding 0.6 and 0.3 times on average), but often stopped short of the goal. We speculate that this might be a result of not observing data similar to the indoor navigation datasets used to train the stronger performing models like VC-1 Base and VC-1 Large.

\textbf{VC-1 Base} and \textbf{VC-1 Large} exhibited contrasting behavior. The larger model explored the environment effectively (high number of steps) and achieved a high success rate, while the base model appeared to randomly explore, but reached the goal when seen from afar.

\subsection{Impact of Model Size, Fine-Tuning, and Data Augmentation}
\label{sec:vc1_variations}

This section studies the impact of three key design decisions when using PVRs for robotics applications: model size, freezing the PVR vs.~fine-tuning, and the use of data augmentations, i.e.~random color jitter and translations added to RGB input during policy learning.
Specifically, our aim is to analyze the impact of each of these variations on simulation predictivity. %
We chose the VC-1 PVR as our basis for investigating these design decisions, which are detailed below:
\begin{compactenum}[\hspace{5pt}1.]
\item \textbf{Model Size}: We used VC-1 models trained on the ViT-Base and ViT-Large architectures (called VC-1 Base and VC-1 Large respectively).
\item \textbf{Freeze vs Fine-tune}: We trained policies for each downstream task under two settings: keeping the PVR frozen (as in the above sections), as well as allowing the weights of the PVR to be fine-tuned by the task objective.
\item \textbf{Data Augmentation}: We trained downstream policies with and without data augmentation. We applied data augmentation in the RGB image space: 20\% random color jitter in brightness, contrast and hue, and 8-pixel random translations.
\end{compactenum}

We trained and evaluated policies for each of 4 downstream tasks with the cross-product of the above variations, both in simulation and in the real world. This produced 32 different tuples of (sim, real) performance. To analyze the impact of each variation, we sliced these tuples along the dimension of each variation and computed the SRCC of the 16 data points in each arm. The results are shown in Figure 5. 
\begin{table}[ht]
\label{fig:vc1vartable}
\centering
\caption{Success rate of policies using two model sizes, with and without fine-tuning and augmentations on 4 tasks.} 
\resizebox{0.48\textwidth}{!}{
\begin{tabular}{c@{\hskip 4pt}ccccccccccccc}
\toprule
& & & & \multicolumn{2}{c}{TriFinger} & \multicolumn{2}{c}{Franka} & \multicolumn{2}{c}{Franka} & \multicolumn{2}{c}{Franka} & \multicolumn{2}{c}{Average} \\
\cmidrule(lr){5-6}
\cmidrule(lr){7-8}
\cmidrule(lr){9-10}
\cmidrule(lr){11-12}
\cmidrule(lr){13-14}
& & & & \multicolumn{2}{c}{Push Cube} & \multicolumn{2}{c}{Reach Pos} & \multicolumn{2}{c}{Pick Up Bottle} & \multicolumn{2}{c}{Open Drawer} & \multicolumn{2}{c}{All tasks}\\

\texttt{\#} & model size & frozen & aug. &   (sim) & (real)  &   (sim) & (real)  &   (sim) & (real) &   (sim) & (real) &   (sim) & (real) \\
\midrule
\texttt{1} & VC-1 Base  & yes & no  & \bf 40 & \bf 37 &      97 &  \bf 97 &       80 &     83 &  \bf 57 &     67 &      55 &      57 \\
\texttt{2} & VC-1 Base  & yes & yes &     38 &     35 &      63 &      40 &       90 &     83 &      40 &     70 &      46 &      46 \\
\texttt{3} & VC-1 Base  & no  & no  &     36 & \bf 37 &      89 &      33 &       93 &     63 &      40 &     53 &      52 &      37 \\
\texttt{4} & VC-1 Base  & no  & yes &     36 & \bf 37 & \bf 100 &      93 &  \bf 100 & \bf 90 &      47 & \bf 73 &  \bf 57 & \bf  59 \\ 
\midrule
\texttt{5} & VC-1 Large & yes & no  & \bf 41 & \bf 38 &      97 &      87 &       77 &     43 &      50 &     57 &      53 &      45 \\
\texttt{6} & VC-1 Large & yes & yes &     35 & \bf 38 &      85 &      37 &  \bf 100 &     60 &      43 &     63 &      53 &      40 \\
\texttt{7} & VC-1 Large & no  & no  &     28 &     34 &      93 &      57 &       97 & \bf 90 &      33 &     43 &      50 &      45 \\
\texttt{8} & VC-1 Large & no  & yes &     34 &     31 &      96 &      87 &  \bf 100 &     50 &      47 &    67  &      55 &      47 \\
\bottomrule
\end{tabular}
}
\end{table}

\begin{table}[ht]
\centering
\caption{Sim2Real transfer results. All results are with policies trained in simulation and evaluated on real robots.}
\label{table:sim2real}
\resizebox{0.48\textwidth}{!}{

\begin{tabular}{c@{\hskip 4pt}ccccccccccccc}
\toprule
& & & & \multicolumn{2}{c}{TriFinger} & \multicolumn{2}{c}{Franka} & \multicolumn{2}{c}{Franka} & \multicolumn{2}{c}{Franka} & \multicolumn{2}{c}{Stretch} \\
\cmidrule(lr){5-6}
\cmidrule(lr){7-8}
\cmidrule(lr){9-10}
\cmidrule(lr){11-12}
\cmidrule(lr){13-14}
& & & & \multicolumn{2}{c}{Push Cube} & \multicolumn{2}{c}{Reach Pos} & \multicolumn{2}{c}{Pick Up Bottle} & \multicolumn{2}{c}{Open Drawer} & \multicolumn{2}{c}{ImageNav}\\
\texttt{\#} & model size & frozen & aug. &   (sim) & (real)  &   (sim) & (real)  &   (sim) & (real) &   (sim) & (real) &   (sim) & (real) \\
\midrule
\texttt{4}  & VC-1 (ViT-B)    & yes & no  & 40    & 3   & 97    & 0     & 80    & 0 & 57    & 23  & 75    & 60 \\
\texttt{5}  & VC-1 (ViT-B)    & yes & yes & 38    & 6   & 63    & 0     & 90    & 0 & 40    & 27  & 75    & 10 \\
\texttt{6}  & VC-1 (ViT-B)    & no  & no  & 36    & 20  & 89    & 0     & 93    & 0 & 40    & 33  & 61    & 60 \\
\texttt{7}  & VC-1 (ViT-B)    & no  & yes & 36    & 11  & 100    & 0     & 100    & 0 & 47    & 30  & 47    & 90 \\
\midrule
\texttt{8} & VC-1 (ViT-L)    & yes & no  & 41    & 2   & 97    & 0     & 77    & 0 & 50    & 23  & 71    & 90 \\
\texttt{9} & VC-1 (ViT-L)    & yes & yes & 35    & 3   & 85    & 0     & 100    & 0 & 43    & 27  & 76    & 80 \\
\texttt{10}  & VC-1 (ViT-L)    & no  & no  & 28    & 23  & 93    & 0     & 97    & 0 & 33    & 37  & 60    & 60 \\
\texttt{11}  & VC-1 (ViT-L)    & no  & yes & 34    &   15  & 96    &   0    & 100    &  0  & 47    &  40   & 69    &  90  \\
\bottomrule
\end{tabular}
}
\end{table}
The SRCC values for all VC-1 variations are statistically significant (p < .05) and show a strong positive correlation, greater than 0.5 in all cases. The differences in \simpred are smaller with regards to the backbone size (0.66 for Base vs. 0.60  for Large) and whether or not the layers were kept frozen (0.58 for frozen vs. 0.64 for fine-tuned). Notably, there is a drop in predictivity from 0.68 to 0.55 when we use augmentations; the reason for this performance drop is that the policies trained with augmentations for the Open Drawer task consistently outperformed their results in simulation when evaluated in the real world, hence affecting predictivity; we hypothesize this performance increase in real world is due to an increase in policy robustness.
As noted, simulation performance is predictive of real world performance to different degrees for all variations (significant SRCC values). From Table IV we can see that, regardless of the size of the backbone; fine-tuning and using augmentations yielded the best results on average and that the model with the highest average performance across all tasks was VC-1 Base with augmentations and fine-tuning. It should be noted that, as with other PVRs, Sim2Real transfer did not work for most tasks even with these variations (Table V). %

\section{CONCLUSIONS}
\label{sec:conclusion}

Our large-scale empirical study has advanced the understanding of pre-trained visual representations (PVRs) in robot learning. We found a high degree of Sim2Real predictivity of PVR-based policies, suggesting that simulation experiments can inform real-world performance. Notably, we have achieved a landmark result on ImageNav, demonstrating the critical role of PVRs in enabling effective Sim2Real transfer. Finally, our study highlights the impact of key design decisions, such as model size, data augmentation, and fine-tuning when deploying PVRs in real-world robotics tasks. These insights help illuminate the potential of PVRs for robot learning, setting a strong foundation for future research.

\section{Acknowledgements}
The Georgia Tech effort was supported in part by ARO PECASE. The views and conclusions contained herein are those of the authors and should not be interpreted as necessarily representing the official policies or endorsements, either expressed or implied, of the U.S. Government, or any sponsor.

\addtolength{\textheight}{-12cm}   %
\endgroup

\bibliographystyle{IEEEtran}
\bibliography{IEEEabrv,references}

\section{APPENDIX}
\begin{figure}%
    \centering
    \includegraphics[width=0.19\columnwidth]{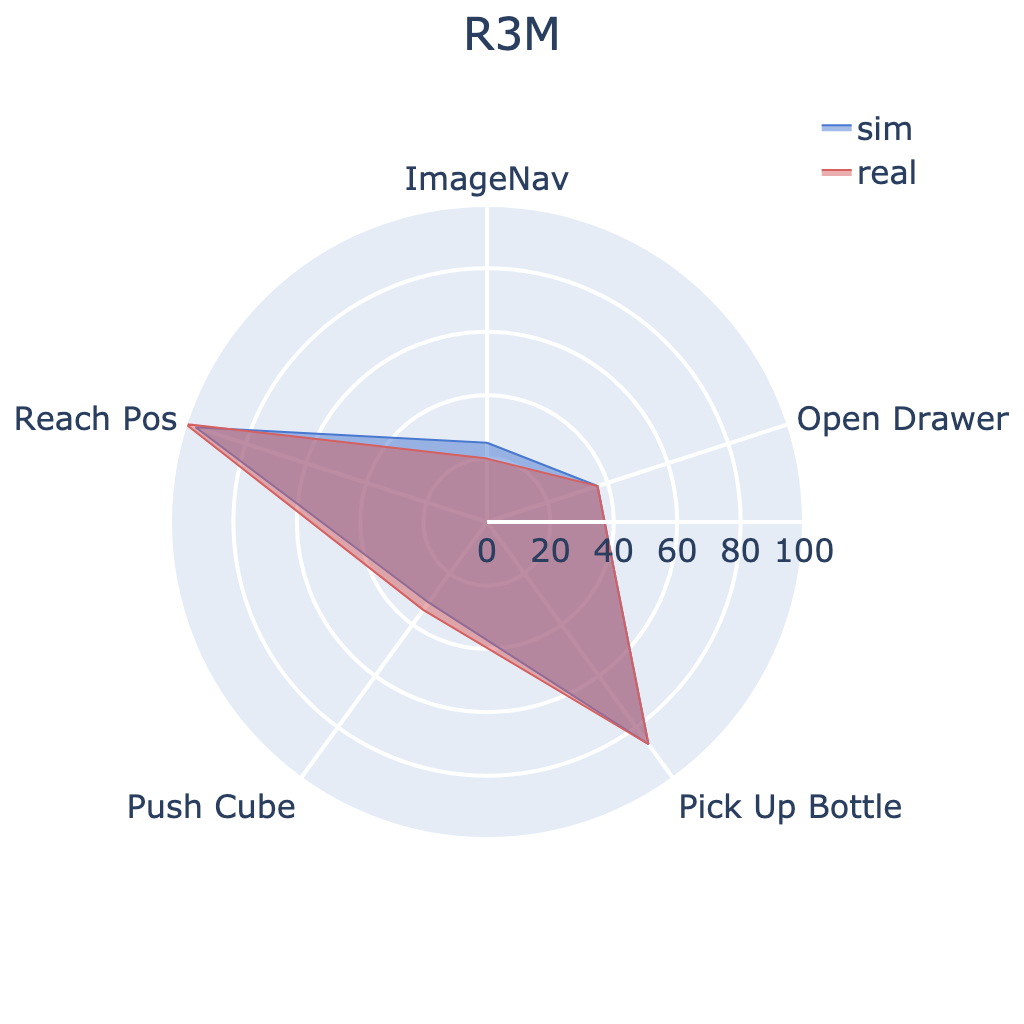}
    \includegraphics[width=0.19\columnwidth]{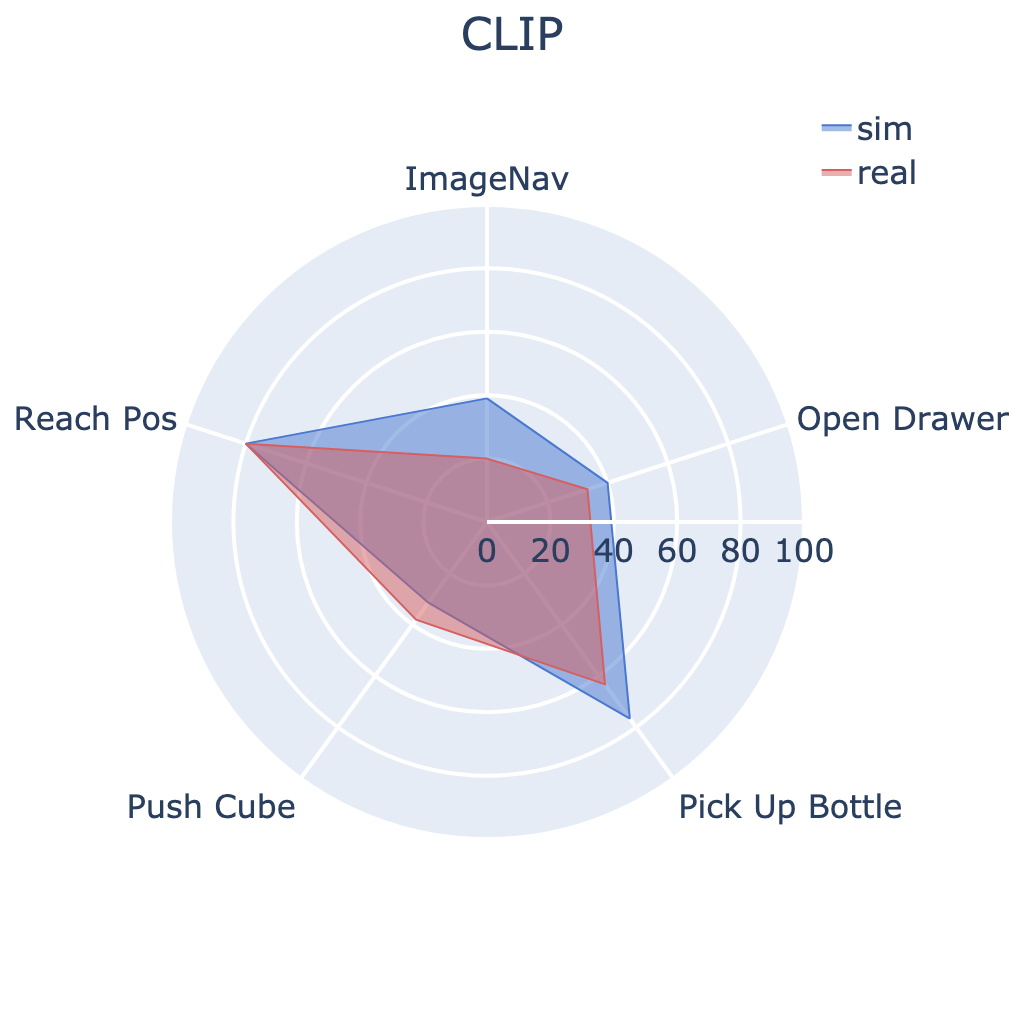}
    \includegraphics[width=0.19\columnwidth]{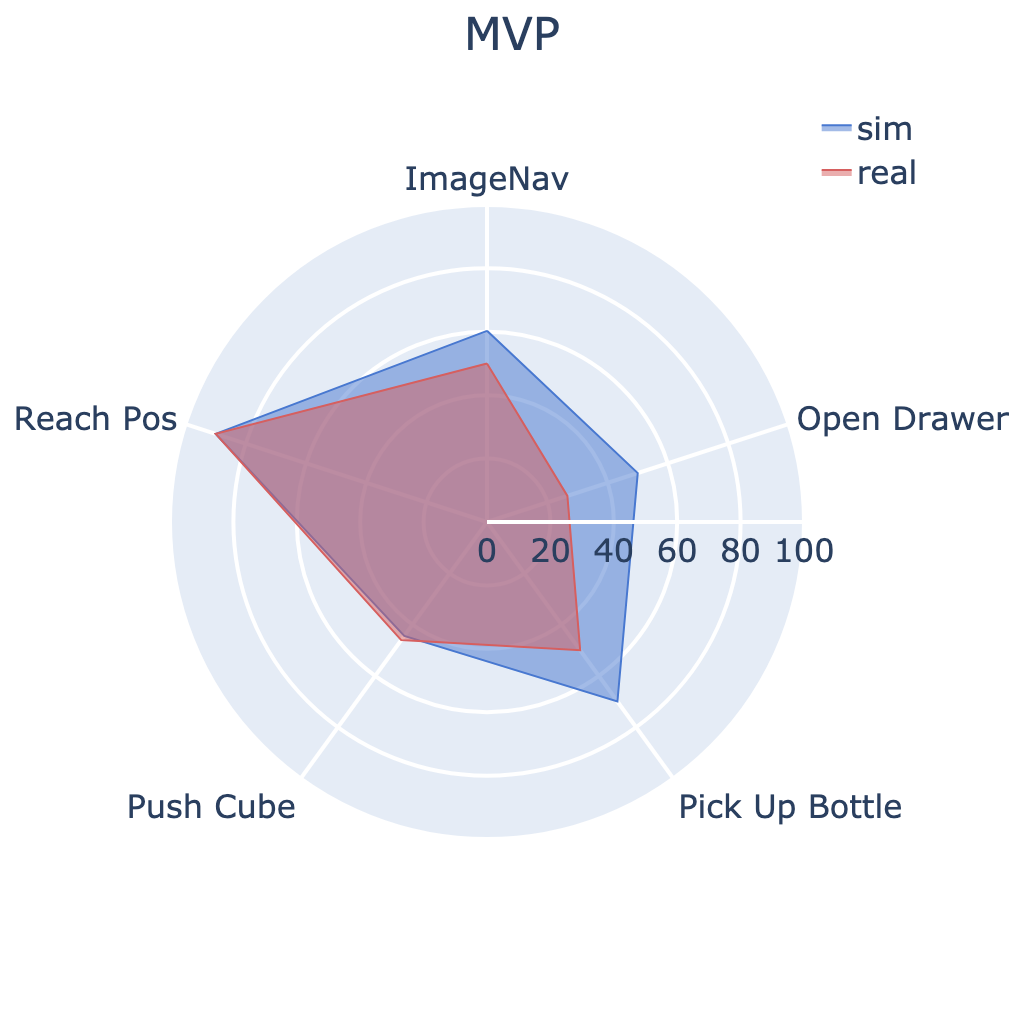}
    \includegraphics[width=0.19\columnwidth]{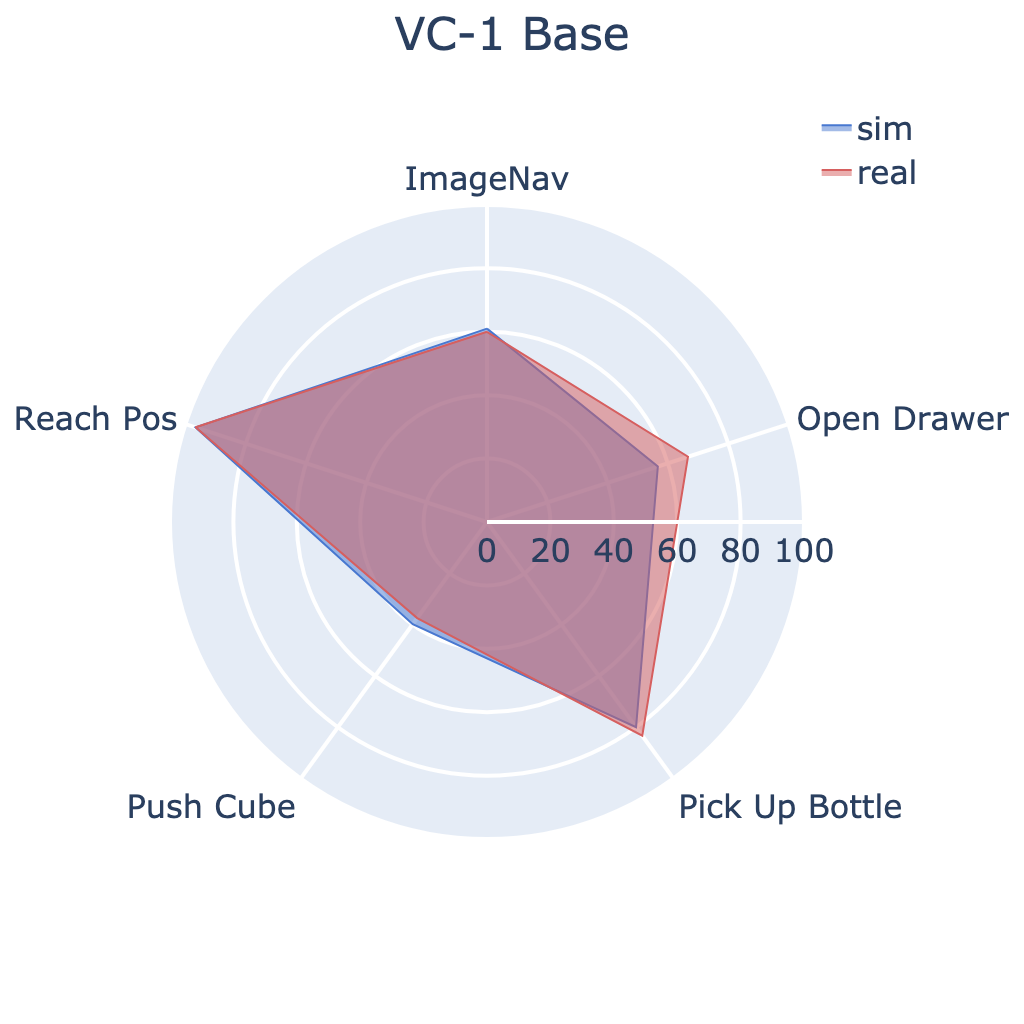}
    \includegraphics[width=0.19\columnwidth]{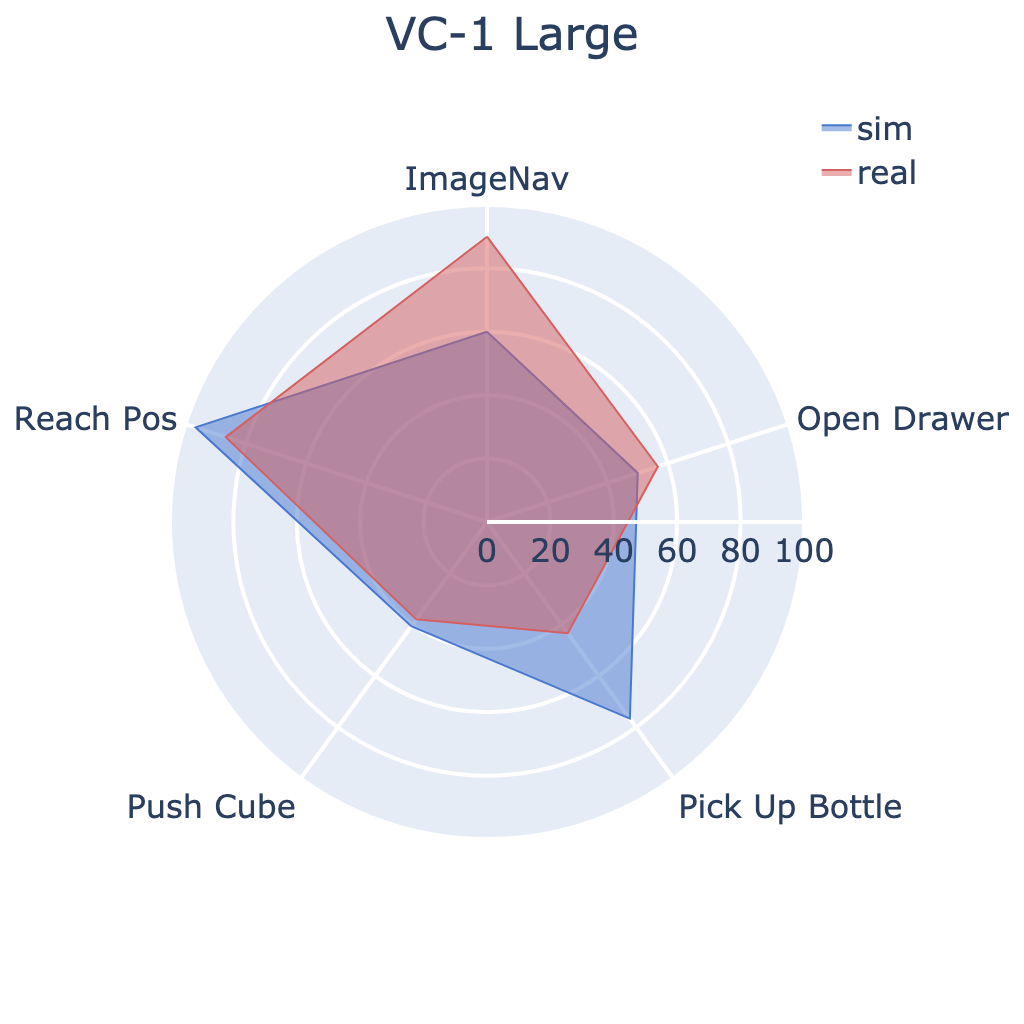}
    \caption{Comparison of performance in simulation (blue) vs. reality (red) for five PVRs on five tasks. R3M~\cite{Nair_r3m_2022} and VC-1 Base~\cite{cortex-vc1-2023} performance in sim closely matches reality on \emph{all} tasks. However, for CLIP~\cite{radford2021learning}, MVP~\cite{Radosavovic_MVP_2022}, and VC-1 Large~\cite{cortex-vc1-2023} there is mismatched performance on multiple tasks. 
    }
    \label{fig:sim_v_real_frozen_pvrs}
\end{figure}

\begin{figure}%
    \centering
    \includegraphics[width=0.89\columnwidth]{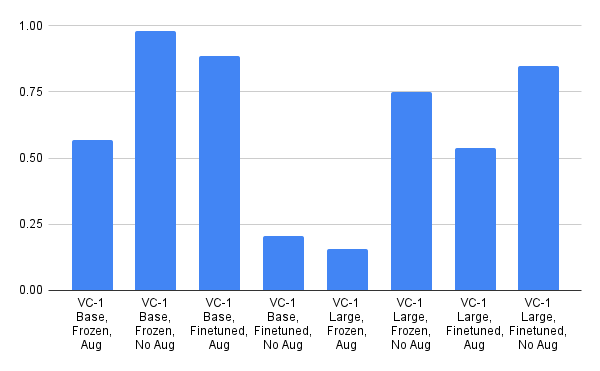}
    \caption{\simpred correlation for each variation of VC-1 Large.}
    \label{fig:srcc_vc1_base_variations}
\end{figure}

\subsection{Details of Manipulation and Navigation tasks}
\label{app:appendix_task_details}
\subsubsection{Task, Policy and Training details for the TriFinger task}
\label{app:appendix_task_details_trifinger}
In this task, the robot must move a cube from an arbitrary initial position to an arbitrary goal position specified by a goal image $I_g$ of the cube at the goal position. The state for the BC policy is $[x_t, z_t, z_g]$, where $x_t$ are the proprioceptive states of the fingers, and $z_t$ and $z_g$ are PVR-encoded versions of the current camera image $I_t$ and the goal image $I_g$, both at a resolution of 270$\times$270. We choose to specify the goal as an image to further underscore the role of visual perception in this task. The initial and goal cube positions are uniformly sampled from within the robot workspace. The success criteria are defined as $1-\frac{d_f}{d_i}$ where $d_f$ and $d_i$ are the final and initial distance between the cube and the goal (respectively).

We collect demonstrations in simulation (PyBullet~\cite{coumans2021}) using a hand-designed expert policy that relies on knowing the ground truth initial cube pose, which is easily obtainable in simulation. Given the initial cube pose, the expert policy first computes the contact points on the cube for each finger (or just one finger in the reach task). It then computes trajectories in Cartesian space for each fingertip from their initial positions to their respective contact points on the cube. Once the fingers have reached the contact points and grasped the cube, the expert policy computes trajectories for each fingertip to bring the cube to the goal position. 
These trajectories are then used to train a policy using behavior cloning, with a learning rate of $10^{-4}$ for 500 epochs. 

We compute the joint torques needed for tracking the desired fingertip trajectories in Cartesian space using the simplified impedance controller from~\cite{Wuthrich_trifinger_2020} (time index omitted for clarity):
\begin{equation} \label{eq:ctr2}
\tau = J^T(k_p(x_{\text{ref}}-x)+k_v(\dot{x}_{\text{ref}}-\dot{x}))
\end{equation}
where $\tau \in\rm I\!R^9$ is the vector of joint torques to be applied to each finger, $x_{\text{ref}}$ are the desired fingertip positions from the reference trajectory, $J$ is the Jacobian of the 3 fingers, and $k_p$ and $k_v$ are hand-tuned controller gains. We also use this controller to execute policies on the real robot.

\begin{figure}[h]
    \centering
    \includegraphics[width=1\columnwidth]{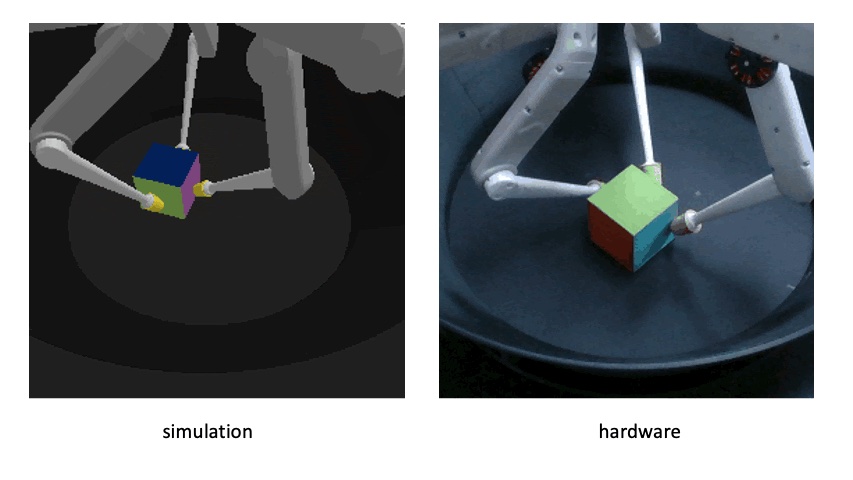}
    \caption{Trifinger Simulation and Hardware Setup}
    \label{fig:trifinger-tasks-images}
    \qquad
\end{figure}

See \cref{fig:trifinger-tasks-images} for a comparison between the sim and real-world visuals for the Trifinger task.

\subsubsection{Task, Policy and Training details for the Franka tasks}
\label{app:appendix_task_details_franka}

\begin{figure}

\centering
\scalebox{1}{
\begin{tabular}{ccc}
\hspace{-12pt}
\subfloat[Reaching Random Point on real robot]{\includegraphics[angle=0,width=0.32\columnwidth]{imgs/franka_reaching_line.jpg}} & \hspace{-12pt}
\subfloat[Bottle Pickup on real robot]{\includegraphics[angle=0,width=0.32\columnwidth]{imgs/franka-pickup.jpg}} & \hspace{-12pt}
\subfloat[Open Drawer task on a kitchen table-top setup on real robot]{\includegraphics[angle=-90,width=0.32\columnwidth]{imgs/franka_open_drawer2.jpeg}} \\
\hspace{-12pt}
\subfloat[Reaching Random Point in sim]{\includegraphics[angle=0,width=0.32\columnwidth]{imgs/franka_reaching_sim.jpg}} & \hspace{-12pt}
\subfloat[Bottle Pickup in sim]{\includegraphics[angle=0,width=0.32\columnwidth]{imgs/franka_pickup_sim.jpg}} & \hspace{-12pt}
\subfloat[Open Drawer task on a kitchen table-top setup in sim]{\includegraphics[angle=0,width=0.32\columnwidth]{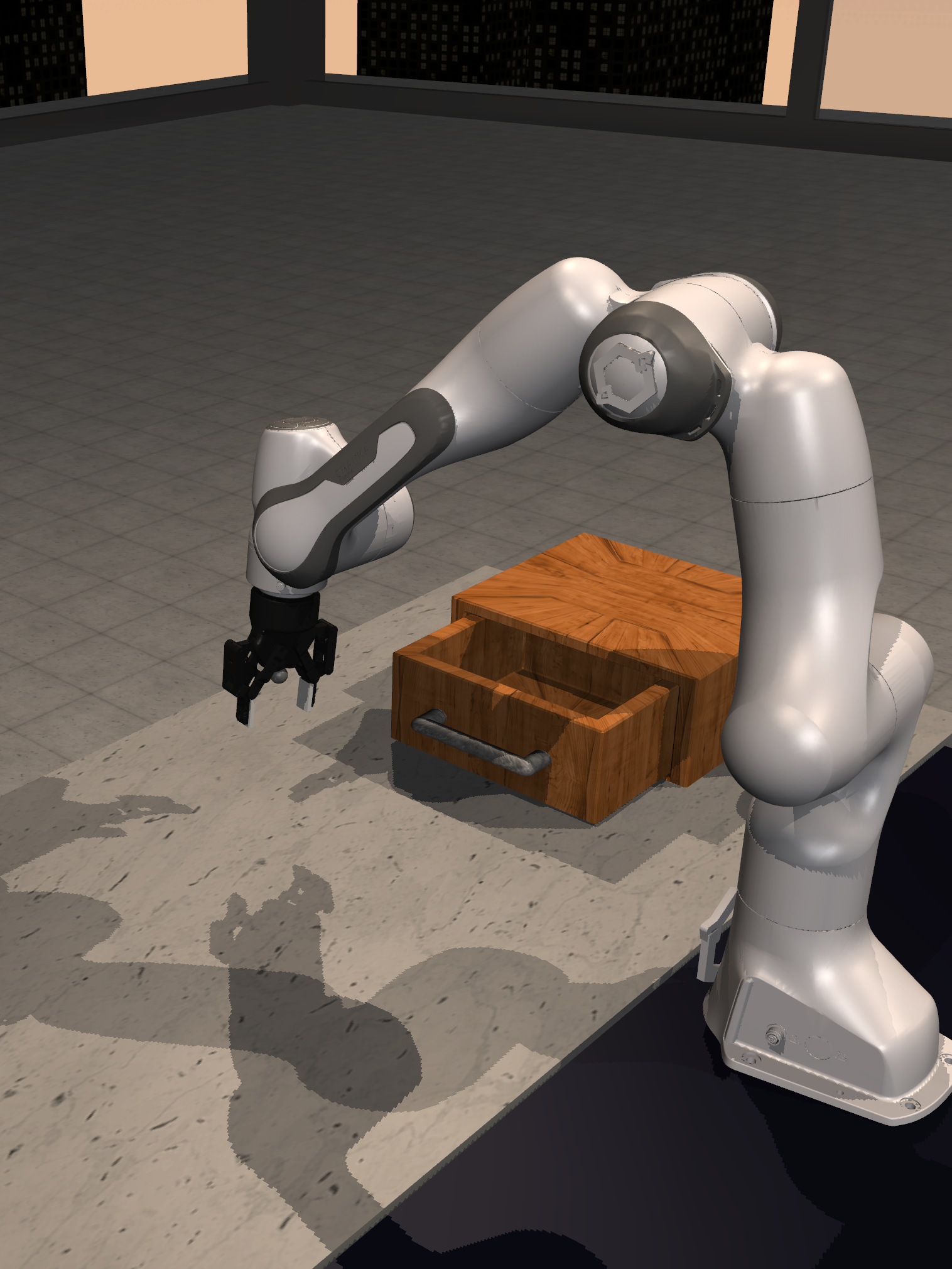}}
\end{tabular}
}
\caption{Franka real-world manipulation tasks}
\label{fig:franka-tasks-images-only}
\end{figure}

\cref{fig:teaser} and \cref{fig:franka-tasks-images-only} show the task configurations, which include reaching a target point with the gripper (Reach Pos) within 5cm error, picking up a bottle stably (Pick Up Bottle), and opening a drawer (Open Drawer) more than 50\% of the way. The target position or object for each task is randomly reset before each demonstration and evaluation so a na\"ive kinesthetic replay does not solve the task. For the Reach Pos task, the target is provided to the policy in the form of a PVR-encoded goal image.
The policies take as input the proprioception of the arm and gripper (joint angles and velocities), and the PVR-encoded 424$\times$240 RGB image taken from a RealSense D435 camera. The learned policies output desired joint angles which are followed by the default PD control loop.
The policies take as input the joint angles and velocities of the arm and gripper, and the PVR-encoded 420$\times$224 RGB image taken from a RealSense D430 camera. The learned policies output desired joint angles which are followed by the default PD control loop.

For the Reach Pos and Pick Up Bottle tasks, we collected 30 demonstrations and trained a behavior cloning policy at a learning rate of $10^{-4}$ for 200 epochs. For the Open Drawer task, we collected 150 demonstrations and trained a behavior cloning policy at a learning rate $10^{-3}$ for $500$ epochs. For finetuning, we used the same learning rate for the policy and PVR encoder. For augmentation, we added 20\% ColorJitter (brightness, contrast, hue) and 8 pixel random shifts to the image.

See \cref{fig:franka-tasks-images-only} for a comparison between the sim and real world visuals for the Franka tasks.

\subsubsection{Task, Policy, and Training details for the ImageNav task}
\label{app:appendix_task_details_stretch}

In Habitat, we represent the Stretch robot as a cylindrical agent with a height of 1.41m and a radius of 0.3m. The Realsense RGBD camera is placed at a height of 1.31m from the ground and aligned vertically, outputting an image of size 640$\times$480 (H$\times$W) with a horizontal field of view of 42$^{\circ}$. We create our own training episode dataset using the HM3D scene dataset~\cite{ramakrishnan2021habitat}, consisting of 800 scenes and 7.2 million training episodes. We allow the agent to take up to 1,000 steps within each episode. The episode is deemed successful if the agent reaches within 1m of the goal position and calls \stopac.
The policy takes as input PVR encodings of the current camera image and the goal image, both downsampled to a resolution of 160$\times$120 and outputs discrete actions.

\begin{figure}[ht]
    \centering
    \subfloat[ImageNav Hardware Setup]{\includegraphics[width=1\columnwidth]{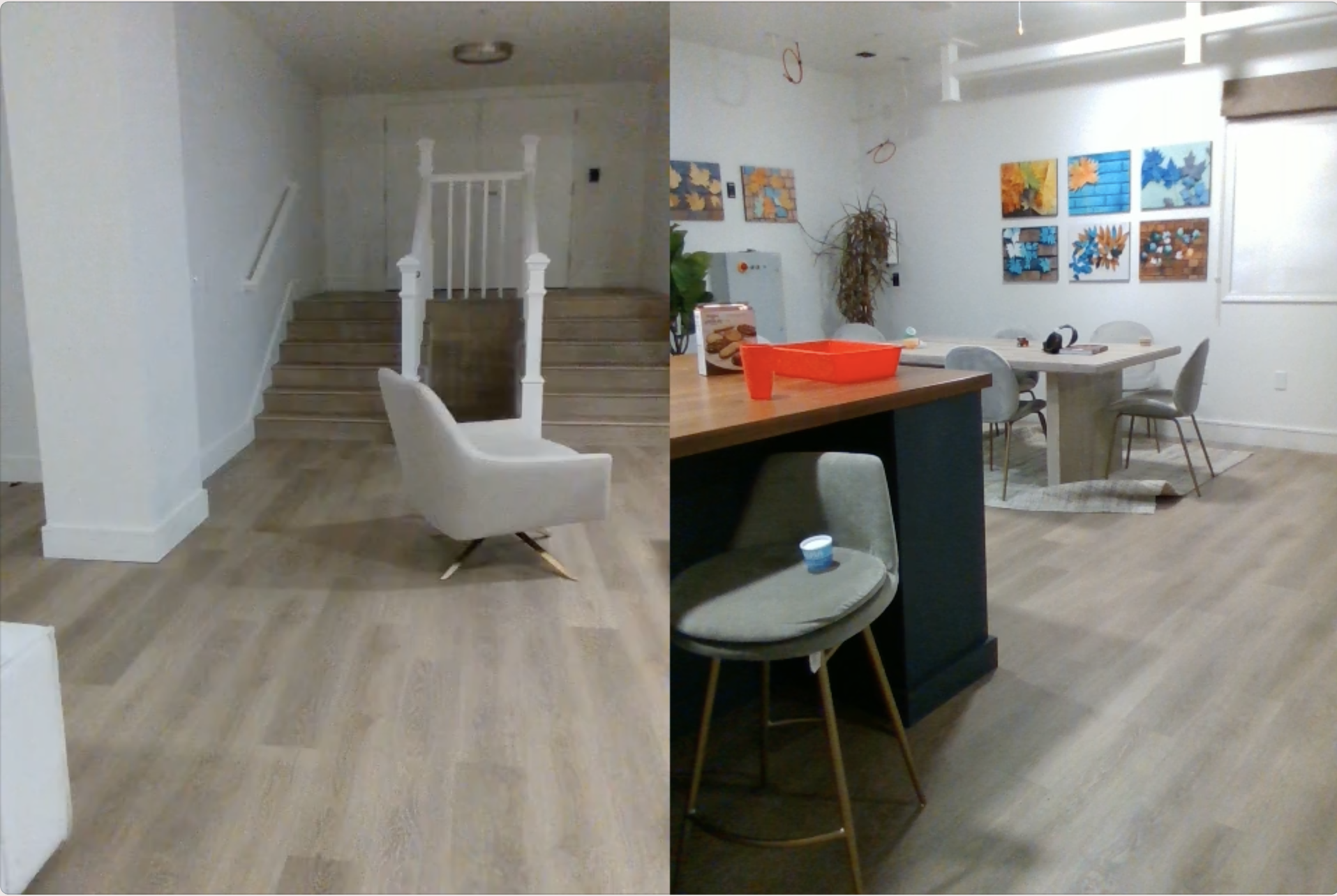}}
    \qquad
    \subfloat[ImageNav Simulation Setup]{\includegraphics[width=1\columnwidth]{imgs/fremont_online.png}}
    \caption{Illustration of the ImageNav task setup for both hardware and simulation platforms.}
    \label{fig:imgnav_setups}
\end{figure}

See \cref{fig:imgnav_setups} for a comparison between the sim and real-world  visuals for the ImageNav task.

\begin{figure}
\centering
\scalebox{1.2}{
\begin{tabular}{cc}
\subfloat[The stretch robot used in the ImageNav real-world experiments.]{\includegraphics[angle=0,height=9cm]{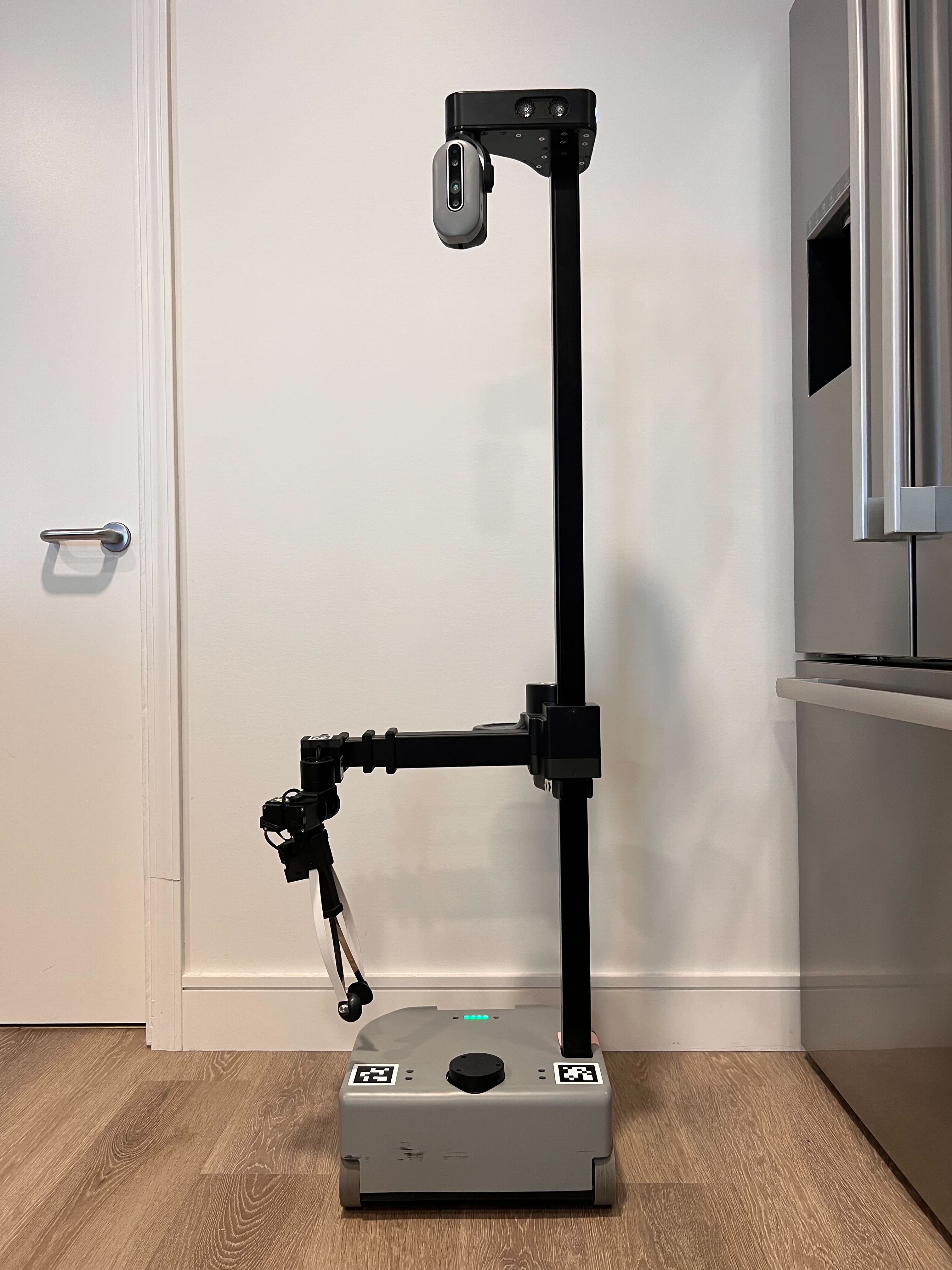}} &
\subfloat[Top-down view of the real-world path (green) the robot took to explore the scene in an episode of the ImageNav task with the point cloud from the robot's head camera and a 2D lidar map.]{\includegraphics[angle=0,height=9cm]{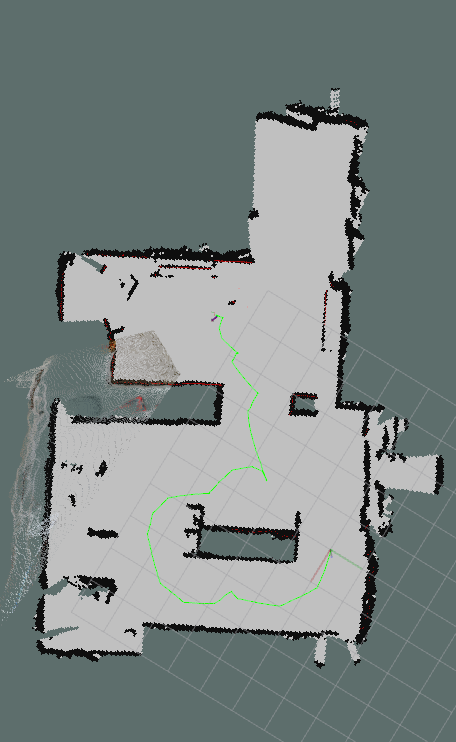}} 
\end{tabular}
}
\caption{Stretch robot and sample of a top-down view of a path from a real-world scene}
\label{fig:stretch-topdown}
\end{figure}

See \cref{fig:stretch-topdown} for a picture of the stretch robot used for our experiments and a top-down map generated by stretch.

\subsection{PVRs details}
\label{app:pvr-details}
\cref{tab:pvr_comparison} is a comparison of the different PVRs studied in this paper.

\begin{table*}[th!]
\caption{Details of the various pre-trained visual representations we evaluate in this work.}
\label{tab:pvr_comparison}
\centering
\resizebox{1.0\textwidth}{!}{
\begin{tabular}{l p{0.4\linewidth} l p{0.4\linewidth} c}
\toprule
Model & Type of supervision & Architecture & Datasets & Number of parameters\\
\midrule
R3M        & Time-contrastive learning, video-text alignment & Resnet50 & Ego4D & 23M \\
CLIP       & Image-text alignment & ViT-B & WebImageText & 86M \\
MVP        & Image only; masked autoencoder & ViT-L & Ego4D, ImageNet, Epic Kitchens, 100DoH, SS & 307M \\
VC-1 Base  & Image only; masked autoencoder & ViT-B & Ego4D, ImageNet, Epic Kitchens, 100DoH, SS-V2, RE10k, OpenHouse24 & 86M \\
VC-1 Large & Image only; masked autoencoder & ViT-L & Ego4D, ImageNet, Epic Kitchens, 100DoH, SS-V2, RE10k, OpenHouse24 & 307M \\
\bottomrule
\end{tabular}
}
\end{table*}

\subsection{Additional experimental data}
\label{app:additional-data}
\cref{tab:frozen_pvrs_success} highlights the differences in performance when models are trained and evaluated in a more realistic setting compared to the original CortexBench. The Franka Open Drawer task is compared with the MetaWorld Open Drawer task, which has a similar specification but uses different robots. The results demonstrate that the success rates reported in CortexBench can be overly optimistic.

\begin{table*}[th   !]
\caption{Comparison of success rates for policies using 5 different frozen PVRs on CortexBench, MetaWorld, and three hardware platforms (TriFinger, Franka, and Stretch) in both simulation and real-world scenarios. }
\label{tab:frozen_pvrs_success}
\centering
\resizebox{1.0\textwidth}{!}{
\begin{tabular}{l@{\hskip 4pt}lccccccccccc} 
\toprule
 & & \multicolumn{1}{l}{CortexBench} & \multicolumn{3}{c}{TriFinger} & \multicolumn{3}{c}{Stretch} & \multicolumn{1}{c}{MetaWorld} & \multicolumn{2}{c}{Franka} \\
 \cmidrule(lr){3-3}
\cmidrule(lr){4-6}
\cmidrule(lr){7-9}
\cmidrule(lr){10-12}
  & Task & \multicolumn{1}{l}{All Tasks} & \multicolumn{3}{c}{Push cube} & \multicolumn{3}{c}{ImageNav} & \multicolumn{3}{c}{Open drawer} \\
\cmidrule(lr){3-3}
\cmidrule(lr){4-6}
\cmidrule(lr){7-9}
\cmidrule(lr){10-12}
\texttt{\#} & Method & CortexBench & CortexBench & sim & real & CortexBench & sim & real & CortexBench & (sim) & (real) \\
\midrule
1 & MVP & 67.5 & 63.4 & 44.4 & 30.0 & 68.1 & 60.3 & 50 & 100.00 & 33.33 & 26.67 \\
2 & R3M & 58.0 & 51.9 & 31.4 & 34.3 & 30.6 & 25.0 & 20 & 100.00 & 37.00 & 36.67 \\
3 & CLIP & 57.0 & 40.1 & 31.5 & 38.1 & 52.2 & 39.0 & 20 & 100.00 & 40.00 & 33.33 \\
4 & VC-1 Base & 66.2 & 60.6 & 39.9 & 37.5 & 67.9 & 61.0 & 60 & 100.00 & 56.67 & 66.67 \\
5 & VC-1 Large & 68.7 & 60.2 & 40.7 & 38.0 & 70.3 & 60.0 & 90 & 100.00 & 50.00 & 56.67 \\
\bottomrule
\end{tabular}
}

\end{table*}

\begin{table*}%
\caption{All experimental results. Dash (-) means that we did not run this particular configuration due to time constraints.}
\label{tab:uber-table}
\resizebox{1.0\textwidth}{!}{
\begin{tabular}{lccccccccccccccc}
\toprule
 &                          Model & Frozen & Augs & \multicolumn{2}{c}{ImageNav} & \multicolumn{2}{c}{Open\_drawer} & \multicolumn{2}{c}{Pick\_up\_bottle} & \multicolumn{2}{c}{Push\_cube} & \multicolumn{2}{c}{Reach\_pos} & \multicolumn{2}{c}{Average} \\
\texttt{\#} &   &&&     real &        sim &         real &         sim &           real &         sim &        real &         sim &         real &          sim &         real &         sim \\
\midrule
0  &           CLIP  &    yes &           no &  20.0±12.65 &  39.0±1.54 &   33.33±3.33 &  40.00±0.00 &    63.33±13.33 &  76.67±3.33 &  38.08±0.38 &  31.48±6.01 &   80.0±11.55 &     80.0±0.0 &   46.95±8.25 &   54.1±2.84 \\
1  &            MVP  &    yes &           no &  50.0±15.81 &  60.3±1.55 &   26.67±3.33 &  50.00±0.00 &      50.0±20.0 &  70.0±11.55 &  30.04±4.08 &   44.4±0.51 &    90.0±5.77 &     90.0±0.0 &    49.34±9.8 &  59.61±3.39 \\
2  &            R3M  &    yes &           no &  20.0±12.65 &  25.0±1.37 &   36.67±3.33 &   37.0±3.33 &     86.67±3.33 &  86.67±8.82 &   34.25±3.1 &  31.37±6.08 &    100.0±0.0 &   96.67±3.33 &   55.52±4.48 &  55.94±5.08 \\
3  &    VC-1 Base &     no &           no &         - &        - &  53.33±13.33 &   57.0±3.33 &     63.33±3.33 &  93.33±3.33 &   37.14±3.7 &  35.75±3.72 &   33.33±6.67 &    88.89±0.0 &   46.78±6.76 &  66.99±4.26 \\
4  &   VC-1 Base  &     no &          yes &         - &        - &   73.33±3.33 &  73.33±6.67 &      90.0±10.0 &   100.0±0.0 &  36.91±1.41 &   35.7±2.86 &   93.33±6.67 &    100.0±0.0 &    73.4±5.35 &  77.26±2.38 \\
5  &   VC-1 Base  &    yes &           no &  60.0±15.49 &  61.0±1.54 &   66.67±3.33 &   70.0±5.77 &    73.33±10.85 &  83.33±3.33 &  37.48±3.43 &   39.85±1.2 &  71.67±11.38 &  70.56±12.12 &    61.83±8.9 &  64.95±4.79 \\
6  &  VC-1 Base  &    yes &          yes &         - &        - &    70.0±5.77 &  73.33±3.33 &     83.33±8.82 &   90.0±5.77 &  34.83±0.46 &  38.05±0.35 &    40.0±10.0 &   62.96±7.41 &   57.04±6.26 &  66.09±4.22 \\
7  &    VC-1 Large  &     no &           no &  60.0±15.49 &  71.0±1.43 &   43.33±3.33 &  50.0±0.00 &      90.0±5.77 &  96.67±3.33 &  34.26±3.27 &  27.94±4.02 &   56.67±6.67 &    92.59±3.7 &   56.85±6.91 &  62.97±4.26 \\
8  &   VC-1 Large   &     no &          yes &   90.0±9.49 &  76.0±1.35 &   66.67±8.82 &  76.67±6.67 &     50.0±25.17 &   100.0±0.0 &   31.0±2.98 &  33.57±0.78 &  86.67±13.33 &     96.3±3.7 &  64.87±11.96 &   76.51±2.5 \\
9  &   VC-1 Large  &    yes &           no &   90.0±9.49 &  60.0±1.55 &   56.67±3.33 &  66.67±6.67 &     48.33±8.72 &  86.67±6.15 &  37.96±1.15 &  40.66±4.55 &   68.33±8.72 &   68.7±13.33 &   60.26±6.28 &  64.54±6.45 \\
10 &  VC-1 Large  &    yes &          yes &  80.0±12.65 &  69.0±1.46 &   63.33±6.67 &   70.0±5.77 &     60.0±20.82 &   100.0±0.0 &  37.93±6.77 &  34.65±2.05 &   36.67±8.82 &    85.19±3.7 &  55.59±11.14 &   71.77±2.6 \\

\bottomrule
\end{tabular}
}
\end{table*}

\begin{table*}%
\caption{All experimental results. Dash (-) means that we did not run this particular configuration due to time constraints.}
\label{tab:sim2real-uber-table}
\resizebox{1.0\textwidth}{!}{
\begin{tabular}{lccccccc}
\toprule
Task & Model & Frozen & Augmentation & {ImageNav}	 & {Open Drawer}	 &  {Pick Up Bottle}	 & {Push Cube}\\
Setting &   &&&      sim &       sim &      sim &   sim \\
\midrule
0	 & 	CLIP	     & 	 no  & 	 no & 	0.20+/-nan(1) & 	0.27+/-0.03(3) & 	0.00+/-0.00(3) & 	0.11+/-0.02(3) \\
1	 & 	MVP	         & 	 no  & 	 no & 	0.50+/-nan(1) & 	0.13+/-0.03(3) & 	0.00+/-0.00(3) & 	0.06+/-0.02(3) \\
2	 & 	R3M	         & 	 no  & 	 no & 	0.20+/-nan(1) & 	0.10+/-0.00(3) & 	0.00+/-0.00(3) & 	0.08+/-0.03(3) \\
3	 & 	VC-1 Base	 & 	no	 & 	no & 	 N/A & 	0.33+/-0.03(3) & 	0.00+/-0.00(3) & 	0.14+/-0.03(3) \\
3	 & 	VC-1 Base	 & 	no	 & 	yes & 	 N/A & 	0.30+/-0.00(3) & 	0.00+/-0.00(3) & 	0.17+/-0.04(3) \\
3	 & 	VC-1 Base	 & 	yes	 & 	no & 	0.60+/-nan(1) & 	0.23+/-0.03(3) & 	0.00+/-0.00(3) & 	0.03+/-0.00(3) \\
3	 & 	VC-1 Base	 & 	yes	 & 	yes & 	 N/A & 	0.27+/-0.07(3) & 	0.00+/-0.00(3) & 	0.04+/-0.01(3) \\
4	 & 	VC-1 Large	 & 	no	 & 	no & 	0.60+/-0.00(2) & 	0.37+/-0.03(3) & 	0.00+/-0.00(3) & 	0.20+/-0.02(3) \\
4	 & 	VC-1 Large	 & 	no	 & 	yes & 	0.90+/-0.00(2) & 	0.33+/-0.03(3) & 	0.00+/-0.00(3) & 	0.14+/-0.02(3) \\
4	 & 	VC-1 Large	 & 	yes	 & 	no & 	0.90+/-nan(1) & 	0.33+/-0.03(3) & 	0.00+/-0.00(3) & 	0.04+/-0.01(3) \\
4	 & 	VC-1 Large	 & 	yes	 & 	yes & 	0.80+/-0.00(2) & 	0.23+/-0.03(3) & 	0.00+/-0.00(3) & 	0.02+/-0.01(3) \\
\bottomrule
\end{tabular}
}
\end{table*}

\begin{landscape}

\centering
\begin{table}[ht]
\centering
\caption{Hyperparameters and tasks setup comparisons between  CortexBench, our simulation and real world settings (Part 1)}
\tiny
\resizebox{1.0\textheight}{!}{
\begin{tabular}{|c|c|c|c|c|c|c|c|c|}
\hline
& \multicolumn{3}{c|}{Trifinger} & \multicolumn{2}{c|}{Stretch } & \multicolumn{2}{c|}{Franka }\\
\hline
Task & \multicolumn{3}{c|}{Push cube} & \multicolumn{2}{c|}{ImageNav } & \multicolumn{2}{c|}{Reach position}\\
\hline
Observation Space & \multicolumn{3}{c|}{RGB + proprio. } & \multicolumn{2}{c|}{RGB} & \multicolumn{2}{c|}{RGB + proprio. }\\
\hline
Action Space & \multicolumn{3}{c|}{Continuous } & \multicolumn{2}{c|}{Discrete } & \multicolumn{2}{c|}{Continuous}\\
\hline
Goal Specification & \multicolumn{3}{c|}{Goal Image} & \multicolumn{2}{c|}{Goal Image} & \multicolumn{2}{c|}{Goal Image }\\
\hline
Policy Learning & \multicolumn{3}{c|}{IL} & \multicolumn{2}{c|}{RL} & \multicolumn{2}{c|}{IL}\\
\hline
Context & CortexBench & sim &real & CortexBench & sim2real & sim & real\\
\hline
Robot & Trifinger & Trifinger & Trifinger & LoCoBot & Stretch & Franka & Franka\\
\hline
Epochs trained & 100 & 500 & 500 & 500m Steps & 600m Step & 200 &\\
\hline
Checkpoint selection & Max eval success & epoch 100 & epoch 100 & Max eval success & Last & epoch 50 & epoch 50\\
\hline
Num demonstrations & 100 & 31 & 31 & - & - & 30 & 30\\
\hline
Number of random seeds & 3 & 3 & 3 & 1 & 1 & 3 & 3\\
\hline
Learning Rate & $10^{-4}$ & $10^{-4}$ & $10^{-4}$ & $2.5 \times 10^{-4}$ & $2.5 \times 10^{-4}$ & $10^{-4}$ & $2.5 \times 10^{-4}$\\
\hline
Augmentations described & - & \multicolumn{2}{c|}{Color jitter + translate } & - & Color jitter + translate & \multicolumn{2}{c|}{Color jitter + translate }\\
\hline
Number of evaluation episodes & - & 12 & 12 & - & 10 & 10 & 10\\
\hline
Sampling of goal position & \multicolumn{3}{c|}{12 different fixed positions} & \multicolumn{2}{c|}{random} & \multicolumn{2}{c|}{random}\\
\hline
Sampling of start position & \multicolumn{3}{c|}{12 different fixed positions} & \multicolumn{2}{c|}{random} & \multicolumn{2}{c|}{random}\\
\hline
Other aspects & \multicolumn{3}{c|}{demos from RL/heuristic policy} & - & - & demos from RL/heuristic policy & demos from tele-operation\\
\hline
\end{tabular}
}
\end{table}

\begin{table}[ht]
\centering
\caption{Comparisons between CortexBench, our simulation and real world settings (Part 2)}
\tiny
\resizebox{1.0\textheight}{!}{
\begin{tabular}{|c|c|c|c|c|c|}
\hline
& \multicolumn{2}{c|}{Franka} & MetaWorld & \multicolumn{2}{c|}{Franka }\\
\hline
Task & \multicolumn{2}{c|}{Pick up bottle } & Open drawer & \multicolumn{2}{c|}{Open Drawer }\\
\hline
Observation Space & \multicolumn{2}{c|}{RGB + proprio. } & RGB + proprio. & \multicolumn{2}{c|}{RGB + proprio. }\\
\hline
Action Space & \multicolumn{2}{c|}{Continuous } & Continuous & \multicolumn{2}{c|}{Continuous }\\
\hline
Goal Specification & \multicolumn{2}{c|}{-} & - & \multicolumn{2}{c|}{-}\\
\hline
Policy Learning & \multicolumn{2}{c|}{IL} & IL & \multicolumn{2}{c|}{IL}\\
\hline
Context & sim & real & CortexBench & sim & real \\
\hline
Robot & Franka & Franka & Sayer & Franka & Franka\\
\hline
Epochs trained & 200 & & 100 & 50 & 50\\
\hline
Checkpoint selection & epoch 50 & epoch 50 & Max eval success & epoch 100 & epoch 100 \\
\hline
Num demonstrations & 30 & 30 & 100 & 150 & 150\\
\hline
Number of random seeds & 3 & 3 & 3 & 3 & 3\\
\hline
Learning Rate & $10^{-4}$ & $10^{-4}$ & - & $10^{-3}$ & $10^{-3}$\\
\hline
Augmentations described & \multicolumn{2}{c|}{Color jitter + translate } & - & \multicolumn{2}{c|}{Color jitter + translate}\\
\hline
Number of evaluation episodes & 10 & 10 & - & 10 & 10\\
\hline
Sampling of goal position & \multicolumn{2}{c|}{random} & & \multicolumn{2}{c|}{random}\\
\hline
Sampling of start position & \multicolumn{2}{c|}{random} & & \multicolumn{2}{c|}{random}\\
\hline
Other aspects & demos from RL/heuristic policy & demos from tele-operation & demos from RL/heuristic policy & demos from RL/heuristic policy & demos from tele-operation \\
\hline
\end{tabular}
}
\end{table}

\end{landscape}

\end{document}